\documentclass[11pt]{article}

\PassOptionsToPackage{table}{xcolor}

\usepackage[final]{acl}
\usepackage{times}
\usepackage{latexsym}
\usepackage[T1]{fontenc}
\usepackage[utf8]{inputenc}
\usepackage{microtype}
\usepackage{inconsolata}

\usepackage{booktabs}
\usepackage{amsmath}
\usepackage{amssymb}
\usepackage{mathtools}
\usepackage{algorithm}
\usepackage{algpseudocode}
\usepackage{graphicx}
\usepackage{pifont}
\usepackage{enumitem}
\usepackage{multirow}
\usepackage{adjustbox}
\usepackage{colortbl}
\usepackage[most]{tcolorbox}
\usepackage{listings}

\definecolor{promptBg}{HTML}{FDF6EC}
\definecolor{promptFrame}{HTML}{D4A76A}
\definecolor{promptTitle}{HTML}{5B3A1E}
\definecolor{codeBg}{HTML}{F0F4F8}
\definecolor{codeFrame}{HTML}{7B9EB8}
\definecolor{codeTitle}{HTML}{2C4A63}

\newtcolorbox{systempromptbox}[1][]{%
  enhanced,
  colback=codeBg, colframe=codeFrame,
  coltitle=white, fonttitle=\bfseries\small,
  colbacktitle=codeFrame,
  boxrule=0.6pt, arc=2.5pt,
  left=6pt, right=6pt, top=4pt, bottom=4pt,
  toptitle=2pt, bottomtitle=2pt,
  #1
}
\newtcolorbox{promptbox}[1][]{%
  enhanced,
  colback=promptBg, colframe=promptFrame,
  coltitle=white, fonttitle=\bfseries\small,
  colbacktitle=promptFrame,
  boxrule=0.6pt, arc=2.5pt,
  left=6pt, right=6pt, top=4pt, bottom=4pt,
  toptitle=2pt, bottomtitle=2pt,
  #1
}

\newcommand{\oai}[1][1em]{\raisebox{-0.2\height}{\includegraphics[height=#1]{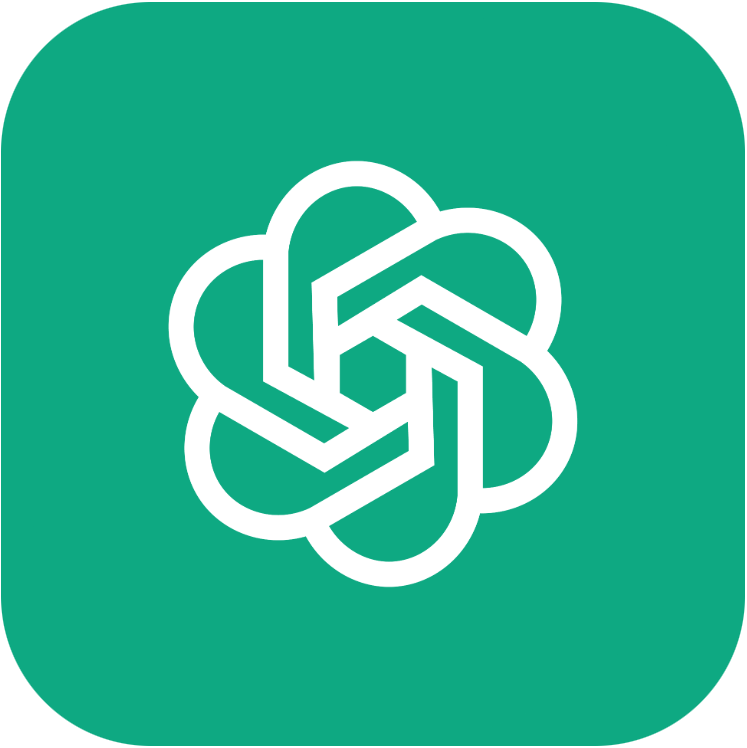}}}
\newcommand{\perplexity}[1][1em]{\raisebox{-0.2\height}{\includegraphics[height=#1]{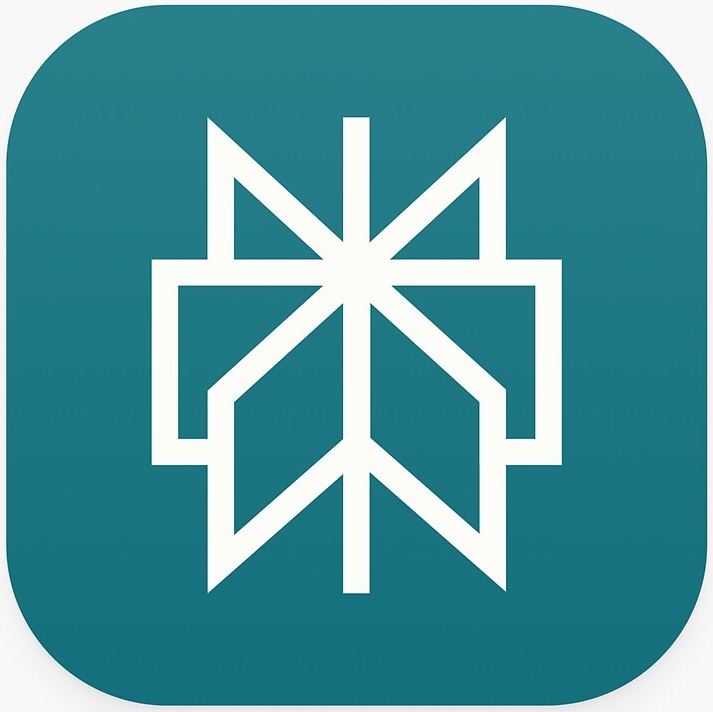}}}
\newcommand{\claude}[1][1em]{\raisebox{-0.2\height}{\includegraphics[height=#1]{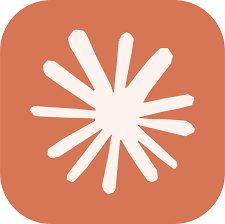}}}
\newcommand{\gemini}[1][1em]{\raisebox{-0.2\height}{\includegraphics[height=#1]{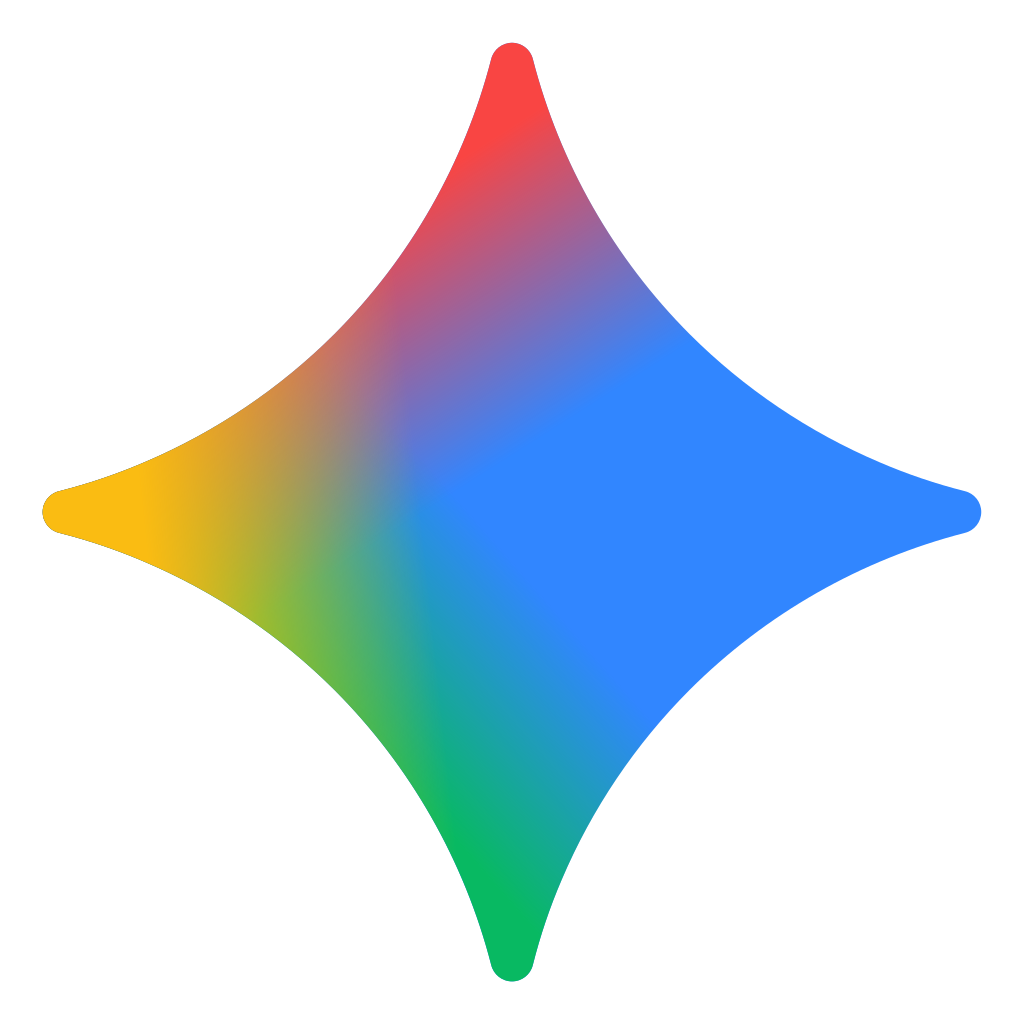}}}
\newcommand{\aitwo}[1][1em]{\raisebox{-0.2\height}{\includegraphics[height=#1]{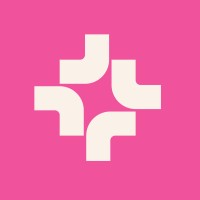}}}
\newcommand{\evtree}[1][1em]{\raisebox{-0.2\height}{\includegraphics[height=#1]{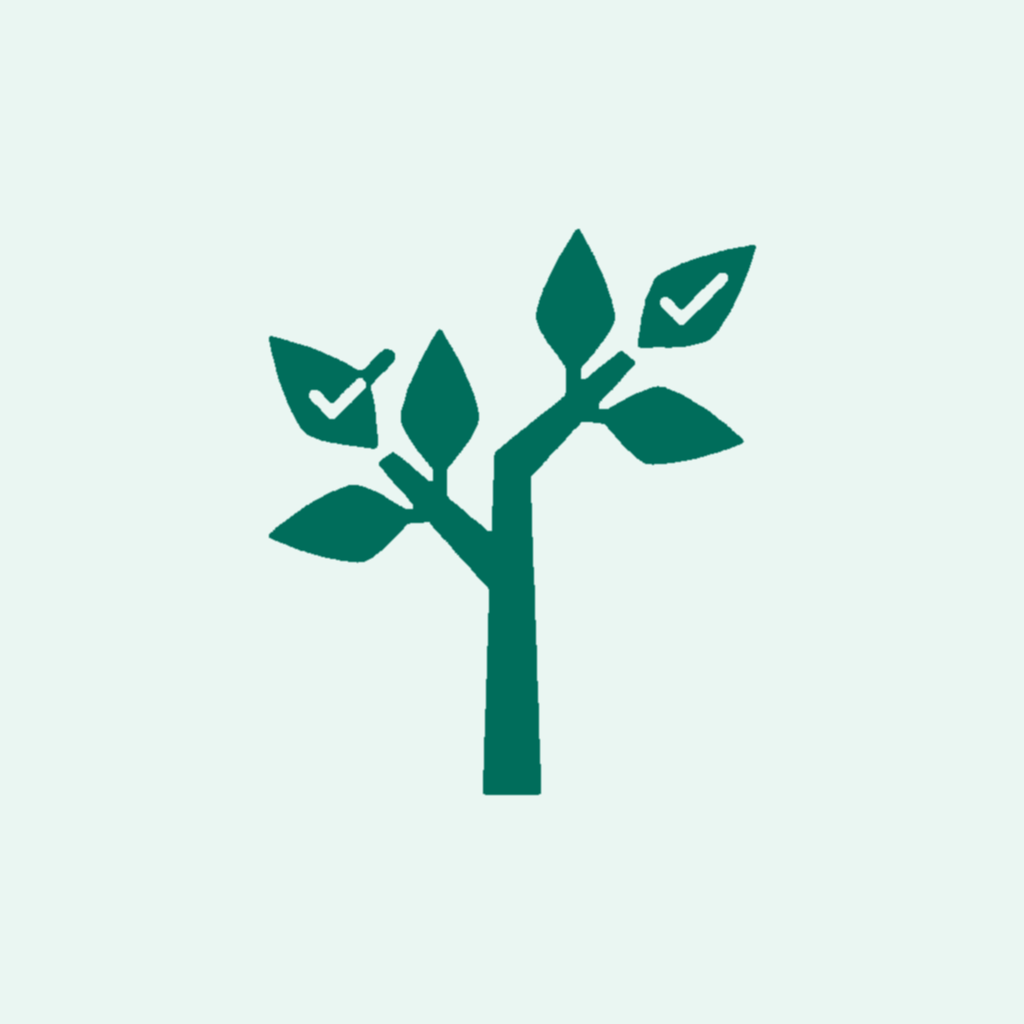}}}

\newcommand{\openmodel}[1][1em]{\raisebox{-0.2\height}{\includegraphics[height=#1]{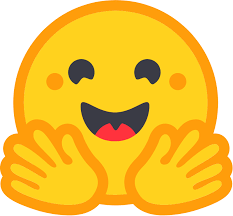}}}
\newcommand{\opencode}[1][1em]{\raisebox{-0.2\height}{\includegraphics[height=#1]{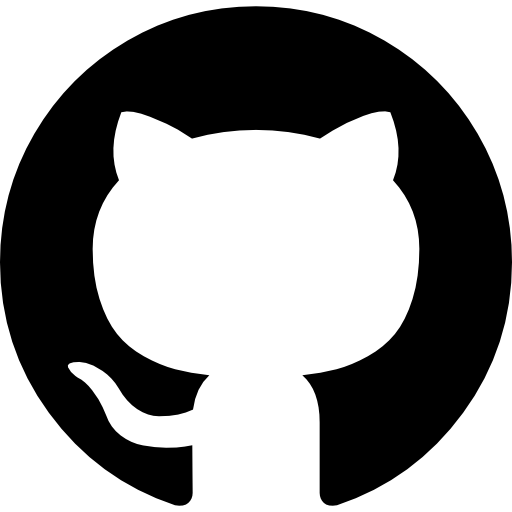}}}

\title{\textsc{DeepRubric}: Evidence-Tree Rubric Supervision for Efficient Reinforcement Learning of Deep Research Agents}

\author{
 \textbf{Minghang Zhu\textsuperscript{1,2}}\space\space\space
 \textbf{Chuyang Wei\textsuperscript{2}}\space\space\space
 \textbf{Junhao Xu\textsuperscript{3}}\space\space\space
 \textbf{Yilin Cheng\textsuperscript{2,3}}\space\space\space
\\
 \textbf{Zhumin Chen\textsuperscript{1}}\footnotemark[1]\space\space\space
 \textbf{Jiyan He\textsuperscript{2}\thanks{Corresponding authors}}
\\
 \textsuperscript{1}Shandong University, Qingdao, China \\
 \textsuperscript{2}Zhongguancun Academy, Beijing, China \\
 \textsuperscript{3}Fudan University, Shanghai, China
\\
\texttt{\detokenize{mhzhu@mail.sdu.edu.cn, hejiyan@zgci.ac.cn}}
}

\begin{document}

\maketitle

\begin{abstract}

Deep research agents synthesize long-form reports by searching and reasoning over retrieved evidence.
Reinforcement learning with rubric-based rewards improves these agents by optimizing them against checkable criteria that translate report quality into reward signals, but its efficiency depends on whether those criteria reliably capture the task scope and evidence needs.
Most existing studies ask an LLM to generate rubrics for a given query, but when the model fails to infer the underlying information needs, the generated rubrics may be incomplete and reduce RL efficiency. 
To obtain more reliable query--rubric supervision, we introduce \textsc{DeepRubric}, a data construction framework that reverses this process: instead of inferring evaluation criteria for a given query, it first determines what an evidence-backed report should be evaluated on and then synthesizes aligned query--rubric pairs from those evaluation targets.
Starting from a sampled seed topic, \textsc{DeepRubric} builds an evidence tree by recursively expanding evidence-backed sub-questions, whose leaves serve as atomic and verifiable evaluation targets.
It then uses the evidence tree to synthesize the training query and rubrics, ensuring that the reward evaluates exactly the information requested by the query.
Using \textsc{DeepRubric}, we construct 9K query--rubric supervision examples and train \textsc{DeepRubric}-8B with rubric-based GRPO, achieving comparable performance to prior open state-of-the-art deep research models across three benchmarks with roughly 13$\times$ fewer RL GPU-hours.
\footnote{Code is available on \url{https://zminghang.github.io/DeepRubric-Code/}.}
\end{abstract}

\section{Introduction}

Deep research agents synthesize long-form reports by searching open corpora and reasoning over retrieved documents~\citep{webthinker,shi2025deep,step_deepresearch}.
Unlike short-form question answering, these agents must identify relevant sources, integrate evidence across documents, support claims with citations, and organize findings into reports that satisfy complex information needs~\citep{researchqa, sharma2025researchrubrics, drb, li2026drb2, deer}.
Recently, reinforcement learning (RL) with rubric-based rewards has become an increasingly common way to improve such systems~\citep{gunjal2025rubricsasrewards, shen2026rrd, drtulu_rler}. In this paradigm, rubrics decompose report quality into checkable criteria, including coverage of important sub-questions, appropriate use of evidence, and citation-supported claims, and then translate them into reward signals to optimize agents~\citep{expertlongbench, viswanathan2025checklists}.
The effectiveness of this training paradigm therefore depends on whether the rubric faithfully captures the task scope and evidence needs behind the research query.

\begin{figure}[!t]
\centering
\includegraphics[width=\columnwidth]{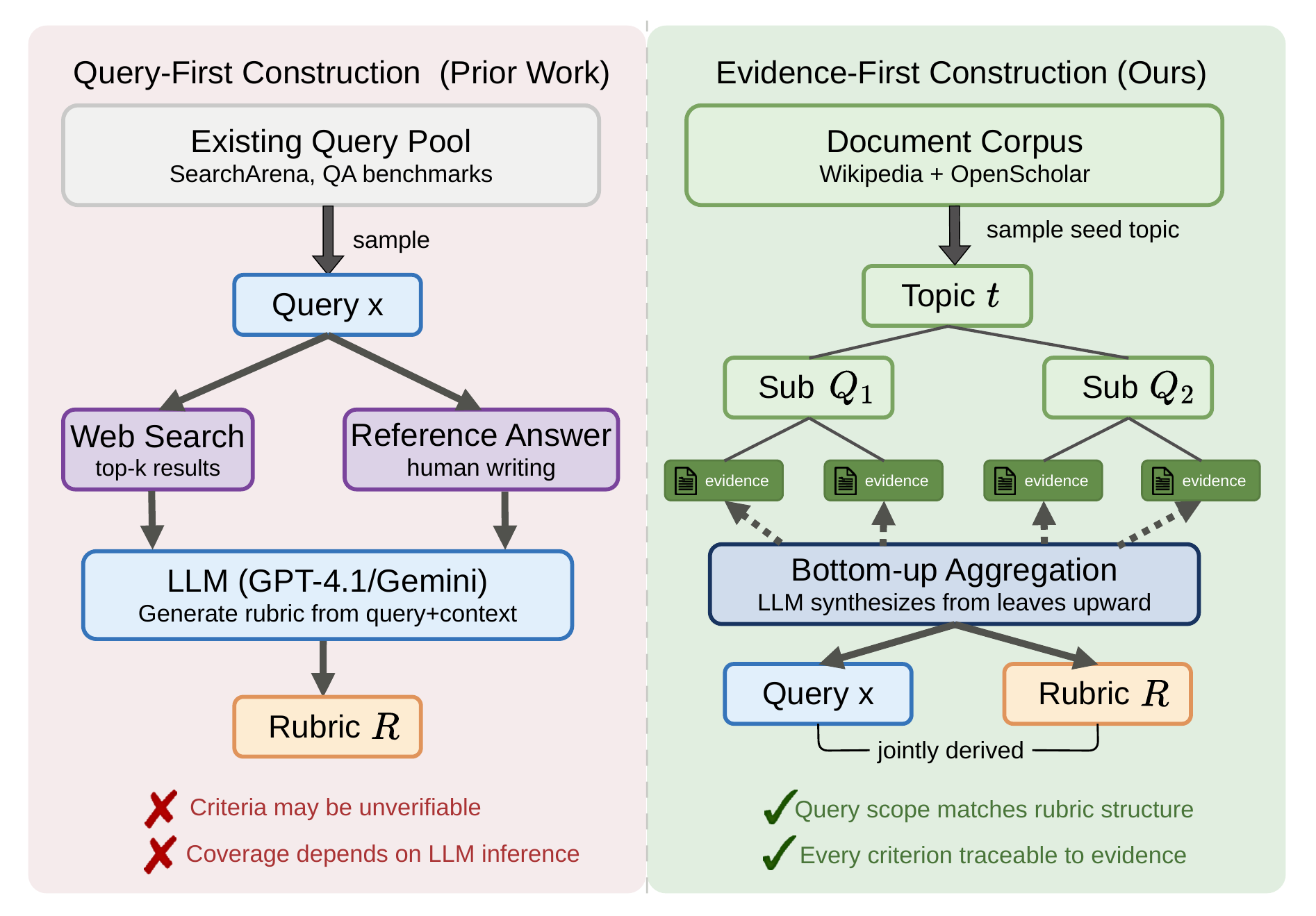}
\caption{\textbf{Query-first vs.\ Evidence-first rubric construction.}
Query-first pipelines infer rubrics from a given query; \textsc{DeepRubric} builds an evidence tree first and derives both the query and rubric from the same  structure, making criteria traceable and query-aligned.}
\label{fig:hero}
\end{figure}

Ensuring such rubric faithfulness is especially important for expensive and time-consuming RL systems, because each rollout may require multi-step tool use and long-form report generation, and prior systems have required thousands to tens of thousands of GPU-hours~\citep{drtulu_rler, asearcher}.
However, most existing pipelines construct rubrics in a query-first manner in Figure~\ref{fig:hero}. Given a user query, they ask a strong language model or human annotator to infer evaluation criteria, often using retrieved documents or reference answers as auxiliary context.
This query-first design is natural, but it is also underdetermined.
A user query rarely reveals the full evidence landscape, and auxiliary context offers only a partial view of what should be evaluated.
As a result, generated rubrics may miss important answerable aspects or include weakly supported criteria, yielding noisy rewards that waste costly rollouts.

To improve the efficiency of deep research RL with more reliable reward supervision, we introduce \textsc{DeepRubric}, a data construction framework that derives training queries and rubric criteria from the same information structure.
Unlike prior pipelines that infer rubrics for a given query, \textsc{DeepRubric} first determines what an evidence-backed report should be evaluated on, and then derives both the rubric and the training query from these evaluation requirements.
Concretely, starting from a sampled seed topic, it recursively expands related sub-questions and grounds them in retrieved documents until reaching fine-grained information needs, which form the leaf nodes of an evidence tree.
These leaf nodes are then converted into atomic rubric criteria, while the training query is synthesized bottom-up from the same tree into a coherent long-form research task.
By deriving the rubric and query from the same tree, \textsc{DeepRubric} makes each criterion traceable to supporting evidence and keeps the generated query aligned with what the reward evaluates.

Using \textsc{DeepRubric}, we construct 9K query--rubric examples from local corpora Wikipedia and OpenScholar without manual expert annotation and train \textsc{DeepRubric}-8B with rubric-based GRPO~\citep{grpo}.
During RL rollouts, the agent retrieves only from the same local corpora used to construct the training tasks, rather than calling external web search APIs.
With only 750 GPU-hours of RL training, \textsc{DeepRubric}-8B matches or surpasses the closest open rubric-RL baseline, DR Tulu-8B~\cite{drtulu_rler}, while using roughly 13$\times$ fewer RL GPU-hours.
Although \textsc{DeepRubric} constructs supervision from local corpora, the learned behavior is not tied to a fixed retrieval environment.
At evaluation time, we replace the local retriever with public online search tools and evaluate on out-of-domain open-retrieval deep research benchmarks, including AstaBench-ScholarQA-CS2~\cite{sqav2}, ResearchQA~\cite{researchqa}, and DeepResearchBench~\cite{drb}.
\textsc{DeepRubric}-8B achieves competitive or superior performance against prior open deep research models, outperforming larger open models with up to 32B parameters, including Tongyi-DR-30B~\cite{tongyi_dr} and WebThinker-32B~\cite{webthinker}.
It also narrows the gap to proprietary systems such as OpenAI Deep Research.


Our contributions are as follows: (i) we introduce \textsc{DeepRubric}, an evidence-first framework for constructing aligned query--rubric supervision from raw corpora; (ii) we train an 8B deep research model with 9K constructed examples, reaching strong open-retrieval performance with only 750 GPU-hours of RL training, and show that doubling the synthesized pool to 18K improves it further; and (iii) we show through ablations, a direct evaluation of the synthesized rubrics, and data analyses that evidence-tree construction, bottom-up query synthesis, and query--rubric alignment are key to the gains.

\section{Related Work}
\label{sec:related_work}

\paragraph{Deep research agents.}
Search-augmented agents are increasingly expected to move beyond short factual answers and produce long-form, evidence-grounded research reports.
Early work mainly studies short-answer or bounded search settings, improving search-augmented reasoning, exploration, and asynchronous RL in interactive environments~\citep{searchr1, deepresearcher, webexplorer, asearcher}.
Recent deep research systems extend this direction to long-form report generation through specialized workflows and report-writing agents~\citep{webthinker, step_deepresearch, agentcpm_report, multimodal_deepresearcher,shi2024generate}, larger multi-tool agents~\citep{tongyi_dr}, and synthetic information-seeking queries or trajectories~\citep{simpledeepsearcher, webdancer, webshaper, drsynth, guo2025peeking}.
These works construct agents, workflows, or behavior traces for producing research reports, but RL still requires training tasks paired with reliable criteria for judging the reports.
\textsc{DeepRubric} focuses on constructing training data for deep research, where each task is paired with rubric-based supervision for long-form answers.

\paragraph{Rubrics for evaluation and RL}
Rubrics provide a practical interface for evaluating open-ended generation by decomposing answer quality into checkable criteria.
They are used in long-form evaluation benchmarks \citep{expertlongbench, sqav2, researchqa, sharma2025researchrubrics, li2026drb2, guo2026rubricreviewer}, and automatic methods generate or refine criteria from task descriptions, retrieved evidence, reference answers, trajectories, or model outputs \citep{shen2026rrd, searchgenv, autorubricr1v, qworld, evalagent, gao2026dr}.
Rubric rewards further enable RL without exact verifiers: Rubrics as Rewards trains with generated criteria \citep{gunjal2025rubricsasrewards}, DR~Tulu evolves search-grounded rubrics from policy rollouts \citep{drtulu_rler}, and related work adapts criteria or critics online \citep{onlinerubrics, rlac, li2026rubricem}.
These criteria generators improve how a given task is judged, but they are largely query-first: they start from a fixed prompt, question, response set, or coarse rubric and infer evaluation criteria afterward.
\textsc{DeepRubric} instead treats evidence-backed information needs as the starting point, building an evidence tree first and co-generating the training query and rubric from the same structure.
\section{Method}
\label{sec:method}

We introduce \textsc{DeepRubric}, an evidence-first pipeline that constructs query-rubric supervision from document corpora for RL post-training of deep research agents.
Given a corpus, \textsc{DeepRubric} first builds an evidence tree by recursively expanding a sampled seed topic into evidence-grounded sub-queries (\S\ref{sec:tree}), then co-synthesizes a training query and verifiable rubric criteria from the same tree (\S\ref{sec:cogen}), and finally uses the rubric as the primary content reward during RL post-training (\S\ref{sec:training}).
We depart from the usual query-first pipeline by first identifying corpus-verifiable information needs and then formulating a query around them, rather than starting from a query and inferring evaluation criteria afterward.

\begin{figure*}[t]
\centering
\includegraphics[width=\textwidth]{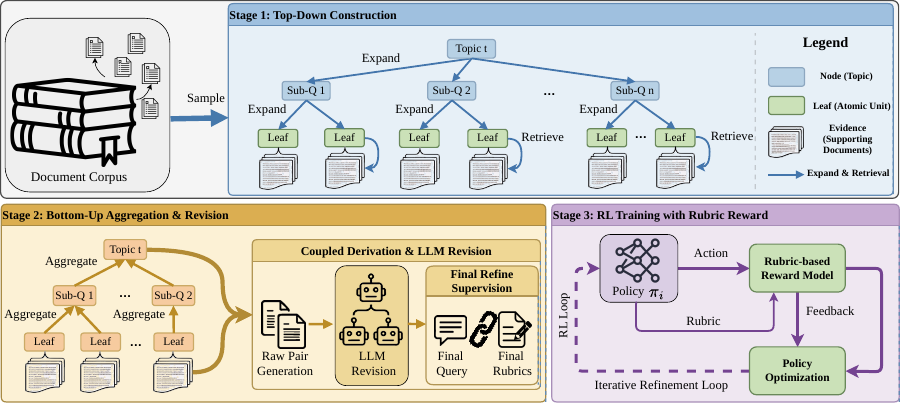}
\caption{\textbf{Method overview.} \textbf{Stage~1}: An evidence tree is constructed top-down from a document corpus; a root topic is recursively decomposed into sub-queries, each grounded in retrieved evidence. \textbf{Stage~2}: The tree is aggregated bottom-up to co-generate a query--rubric pair $(x, \mathcal{R}(\mathcal{T}))$; a verification-and-revision step checks evidence support, query--rubric alignment, and criterion quality. \textbf{Stage~3}: The policy $\pi_\theta$ generates reports conditioned on the query, a rubric-based reward model scores them against $\mathcal{R}(\mathcal{T})$, and the policy is updated via GRPO.}
\label{fig:method_pipeline}
\end{figure*}

\subsection{Evidence Tree Construction}
\label{sec:tree}

\paragraph{Task representation.}
We represent each training task as a rooted evidence tree $\mathcal{T} = (\mathcal{V}, \mathcal{E}, v_0)$ with root node $v_0$.
Each node $v \in \mathcal{V}$ contains three components: (1)~a sub-query $q_v$ that specifies what information this node seeks; (2)~a set of retrieved documents $\mathcal{P}_v = \{p_{v,1}, \ldots, p_{v,k_v}\}$ that provide evidence relevant to $q_v$; and (3)~a set of children $\operatorname{Ch}(v)$, each representing a finer-grained sub-query derived from $q_v$.
A node is called a \emph{leaf} if it has no children: $\mathcal{L}(\mathcal{T}) = \{v \in \mathcal{V} : \operatorname{Ch}(v) = \emptyset\}$.
The tree is a recursive decomposition of a broad topic (the root $v_0$) into progressively more specific, evidence-grounded sub-queries.

\paragraph{Root initialization.}
Given a corpus $\mathcal{C}$ and retriever $\rho$, we sample a seed topic $t$ from $\mathcal{C}$, convert it into a broad research sub-query $q_{v_0}$, and retrieve documents $\mathcal{P}_{v_0} \leftarrow \rho(q_{v_0})$.

\paragraph{Top-down expansion.}
The tree is built by breadth-first expansion.
For each frontier node $v$, an LLM $\mathcal{M}$ observes its sub-query $q_v$, retrieved documents $\mathcal{P}_v$, and ancestor path $\operatorname{Anc}(v)$, and proposes complementary child sub-queries that decompose $q_v$ into finer-grained aspects:
\begin{equation}
\label{eq:decompose}
\widetilde{\mathcal{Q}}_v
= \mathcal{M}\bigl(q_v, \mathcal{P}_v, \operatorname{Anc}(v)\bigr).
\end{equation}
For each proposed child sub-query $\tilde{q} \in \widetilde{\mathcal{Q}}_v$, we retrieve its document set $\tilde{\mathcal{P}}=\rho(\tilde{q})$ and ask $\mathcal{M}$ to ground the child node by selecting the documents that support answering $\tilde{q}$. The resulting grounded pair $(\tilde{q}, \tilde{\mathcal{P}})$ is added as a child of $v$. If $\mathcal{M}$ cannot propose further meaningful child sub-queries, or if the proposed children cannot be grounded in retrieved evidence, $v$ becomes a leaf.

To prevent unbounded growth, we impose structural budgets on the evidence tree: a maximum depth of 3, with depth-aware branching that allows up to 6 children at the root, 4 at intermediate levels, and 3 at deeper levels.
The output of this stage is an evidence tree $\mathcal{T}$, a hierarchical decomposition of the topic into progressively finer sub-queries, each grounded in retrieved documents.
Further details are provided in Appendix~\ref{app:tree_pipeline} and Algorithm~\ref{alg:pipeline}.

\subsection{Query-Rubric Co-Generation and Revision}
\label{sec:cogen}
\label{sec:revision}
The tree $\mathcal{T}$ captures the information structure of a topic, but training an RL agent requires a concrete query-rubric pair $(x, \mathcal{R}(\mathcal{T}))$. We co-generate both by presenting $\mathcal{T}$ to an LLM $\mathcal{M}$, which synthesizes them in a bottom-up pass guided by the tree, then verify the result with a separate model. 

\paragraph{Bottom-up synthesis.}
Given the tree $\mathcal{T}$, we prompt $\mathcal{M}$ to reason from the leaves upward, progressively merging leaf-level content into more abstract summaries at each internal node (the full prompt is provided in Appendix~\ref{app:tree_pipeline}). At the root, the model produces two outputs: (1)~a natural research query $x$ that integrates the information captured by the tree, and (2)~a set of rubric criteria $\mathcal{R}(\mathcal{T})$ derived from the leaf content. The merge process makes the query's scope follow the evidence structure of $\mathcal{T}$ rather than being generated independently.

\paragraph{Rubric structure.}
Each criterion is a typed, weighted, evidence-grounded check:
\begin{equation}
\label{eq:rubric}
r = \bigl(c_r, \mathcal{P}_r, \tau_r, w_r\bigr), \qquad r \in \mathcal{R}(\mathcal{T}),
\end{equation}
where $c_r$ is a natural-language verifiable criterion, $\tau_r \in \{\textsc{factual}, \textsc{logical}\}$ is its type, and $w_r \in [0,1]$ is its importance weight.
For \textsc{factual} criteria, $\mathcal{P}_r$ contains supporting  documents from selected leaves and the report should state information supported by those documents.
For \textsc{logical} criteria, $\mathcal{P}_r=\emptyset$ because the criterion evaluates synthesis, comparison, or reasoning over the report rather than the mention of a specific evidence fact.
The model may discard peripheral or redundant leaves during synthesis, so $\mathcal{R}(\mathcal{T})$ selects the criteria that jointly define a well-scoped, answerable research task.

\paragraph{Verification and revision.}
To improve the consistency of the synthesized supervision, we apply a separate verifier $\mathcal{M}_{\mathrm{verify}}$ to check each tuple $(x, \mathcal{T}, \mathcal{R}(\mathcal{T}))$:
\begin{equation}
\label{eq:verify}
\bigl(d, x', \mathcal{R}'\bigr)
= \mathcal{M}_{\mathrm{verify}}\bigl(x, T, \mathcal{R}(\mathcal{T})\bigr),
\end{equation}
where $d \in \{\textsc{Keep}, \textsc{Revise}, \textsc{Drop}\}$ denotes the verifier decision.
The verifier checks evidence support, query--rubric scope alignment, and criterion quality, including atomicity and redundancy.
For \textsc{Keep} examples, we retain the original tuple; for \textsc{Revise} examples, we use the corrected query $x'$ and rubric $\mathcal{R}'$; and for \textsc{Drop} examples, we remove the tuple from training.
We provide the full verification prompt together with an analysis of revision and drop decisions in Appendix~\ref{app:data_stats} and \ref{app:verify_stats}.

\subsection{RL with Rubric Rewards}
\label{sec:training}

We use the verified query--rubric pairs as reward supervision for RL post-training.
Given a training query $x$, the policy $\pi_\theta$ samples reports $y \sim \pi_\theta(\cdot \mid x)$.
We use standard GRPO for optimization, and the reward is computed by applying the synthesized rubric to each sampled report $y$.

\paragraph{Rubric reward.}
For each criterion $r \in \mathcal{R}(\mathcal{T})$, we use LLM-as-a-judge to score whether the generated report $y$ satisfies the criterion statement $c_r$.
The score is assigned on a 0-4 integer scale and normalized to $s_r(y)\in[0,1]$.
The full scoring prompt is provided in Appendix~\ref{app:rubric_scoring}.
The rubric reward is the importance-weighted mean:
\begin{equation}
\label{eq:rubric_reward}
R_{\mathrm{rubric}}(y) = \frac{\sum_{r \in \mathcal{R}(\mathcal{T})} w_r \cdot s_r(y)}{\sum_{r \in \mathcal{R}(\mathcal{T})} w_r},
\end{equation}
where $s_r(y) \in [0,1]$ is the normalized judge score for criterion $r$.
\paragraph{Composite reward.}
Following prior deep-research RL training~\citep{drtulu_rler}, we combine the rubric reward with auxiliary rewards for format, citation quality, and tool use:
\begin{equation}
\label{eq:total_reward}
\begin{split}
R(y) ={}& 0.5 \cdot R_{\mathrm{rubric}}(y)
 + 0.2 \cdot R_{\mathrm{format}}(y) \\
& + 0.2 \cdot R_{\mathrm{cite}}(y)
 + 0.1 \cdot R_{\mathrm{search}}(y),
\end{split}
\end{equation}
where $R_{\mathrm{format}}$ rewards adherence to the expected output structure (\texttt{<think>}, \texttt{<tool\_call>}, \texttt{<answer>} tags), $R_{\mathrm{cite}}$ measures per-claim citation quality over cited sources, and $R_{\mathrm{search}}$ encourages sufficient use of retrieval tools during generation.
The rubric reward receives the largest weight because it provides the main signal for evidence-grounded content synthesis.
Full reward definitions and details are provided in Appendix~\ref{app:reward_details}.

\paragraph{Optimization.}
We optimize the policy with Group Relative Policy Optimization (GRPO; \citealp{grpo}), sampling a group of candidate reports for each query and computing advantages relative to the group. 
Model initialization, training size, and hyperparameters are reported in Appendix~\ref{app:hyperparams}.

\section{Experiments}
\label{sec:experiments}

\definecolor{oursRow}{HTML}{E8F5E9}     
\definecolor{ablRow}{HTML}{FFF3E0}      
\definecolor{closedRow}{gray}{0.92}     
\definecolor{headerRow}{gray}{0.95}     
\definecolor{ForestGreen}{HTML}{2E7D32} 

We evaluate \textsc{DeepRubric} from three perspectives.
First, we test whether evidence-tree supervision improves open-retrieval deep research performance (\S\ref{sec:main_results}).
Second, we measure how efficiently the model reaches competitive performance under RL (\S\ref{sec:training_efficiency}).
Third, we analyze which construction choices affect supervision quality through ablations, supervision scaling and training-data analysis (\S\ref{sec:ablation}--\S\ref{sec:case_study}).

\subsection{Experimental Settings}
\label{sec:exp_setup}

\noindent\textbf{Datasets and Evaluation Metrics.}
We evaluate deep research agents on three long-form, open-retrieval benchmarks spanning scientific and general research domains: \textbf{AstaBench-ScholarQA-CS2} (SQAv2; \citealp{sqav2}), which assesses synthesis over up-to-date scientific literature; \textbf{ResearchQA}~\citep{researchqa}, which covers academic research questions across diverse fields; and \textbf{DeepResearch Bench} (DRB; \citealp{drb}), which evaluates open-ended research tasks requiring comprehensive reports with citations.
All benchmarks follow their official evaluation protocols.  The details are provided in Appendix~\ref{app:benchmarks}.

\begin{table*}[t]
\centering
\scriptsize
\renewcommand{\arraystretch}{1.15}
\setlength{\tabcolsep}{2.5pt}
\resizebox{\textwidth}{!}{%
\begin{tabular}{@{}ll*{12}{c}@{}}
\toprule
\multirow{2}{*}{\textbf{Model}} 
& \multirow{2}{*}{\textbf{Backbone}} 
& \multicolumn{5}{c}{\textbf{SQAv2}} 
& \multirow{2}{*}{\textbf{ResearchQA}} 
& \multicolumn{5}{c}{\textbf{DRB}} 
& \multirow{2}{*}{\textbf{Average}} \\
\cmidrule(lr){3-7}\cmidrule(lr){9-13}
& & \textbf{Overall} & \textbf{Rubric} & \textbf{Answer} & \textbf{Cite-P} & \textbf{Cite-R}
& & \textbf{Overall} & \textbf{Comp.} & \textbf{Depth} & \textbf{Instr.} & \textbf{Read.}
& \\
\midrule
\multicolumn{14}{@{}l}{\cellcolor{headerRow}\textbf{\textit{Proprietary Deep Research}}} \\
\rowcolor{closedRow} \claude\ Claude-Sonnet Search & Claude Sonnet & -- & -- & -- & -- & -- & 64.3$^*$ & 34.5$^*$ & 39.0 & 37.7 & 45.8 & 41.5 & -- \\
\rowcolor{closedRow} \perplexity\ Perplexity-Sonar (High) & Proprietary & -- & -- & -- & -- & -- & 69.1$^*$ & 40.7$^*$ & 37.4 & 36.1 & 45.7 & 44.7 & -- \\
\rowcolor{closedRow} \perplexity\ Perplexity Deep Research & Proprietary & 67.3 & 91.6 & 92.7 & 47.3 & 37.6 & 75.3$^*$ & 42.3$^*$ & 40.7 & 39.3 & 46.4 & 44.3 & 61.6 \\
\rowcolor{closedRow} \gemini\ Gemini Deep Research & Proprietary & -- & -- & -- & -- & -- & 68.5$^*$ & 48.8$^*$ & 48.5 & 48.5 & 49.2 & 49.4 & -- \\
\rowcolor{closedRow} \gemini\ Gemini 3 Pro + Search & Gemini 3 Pro & 69.8 & 83.1 & 98.3 & 68.5 & 29.4 & 74.3 & 46.3 & 43.4 & 44.9 & 49.8 & 49.0 & 63.5 \\
\rowcolor{closedRow} \oai\ GPT-5 + Search & GPT-5 & 74.8 & 92.3 & 93.8 & 67.8 & 45.6 & 78.2$^\dagger$ & 50.7 & 49.7 & 51.5 & 51.6 & 48.5 & 67.9 \\
\rowcolor{closedRow} \oai\ OpenAI Deep Research & o3/o4 & 79.6 & 91.5 & 95.6 & 77.4 & 43.1 & 79.2$^\dagger$ & 46.9$^*$ & 46.8 & 45.2 & 49.2 & 47.1 & 68.6 \\
\midrule
\multicolumn{14}{@{}l}{\cellcolor{headerRow}\textbf{\textit{Naive RAG}}} \\
Qwen3-8B + RAG & Qwen3-8B & 40.4 & 69.2 & 92.3 & -- & -- & 56.1 & 33.3 & 29.4 & 27.0 & 40.2 & 41.1 & 43.3 \\
QwQ-32B + RAG & QwQ-32B & 41.9 & 77.5 & 90.3 & -- & -- & 60.9 & 40.3 & 38.1 & 34.8 & 47.0 & 44.6 & 47.7 \\
\midrule
\multicolumn{14}{@{}l}{\cellcolor{headerRow}\textbf{\textit{Open Deep Research Models}}} \\
Search-R1-7B \openmodel\ \opencode & Qwen2.5-7B & 22.2 & 9.7 & 79.0 & -- & -- & 27.9 & 9.5 & 5.2 & 2.1 & 18.6 & 16.8 & 19.9 \\
ASearcher-Web-7B \openmodel\ \opencode & Qwen2.5-7B & 26.9 & 13.7 & 94.0 & -- & -- & 19.4 & 7.8 & 5.1 & 1.7 & 15.2 & 11.8 & 18.0 \\
WebExplorer-8B & Qwen2.5-8B & 42.5 & 78.6 & 91.4 & -- & -- & 64.8 & 36.7 & 33.7 & 28.5 & 45.7 & 42.2 & 48.0 \\
WebThinker-32B-DPO & Qwen2.5-32B & 32.9 & 36.7 & 94.9 & -- & -- & 48.6 & 23.3 & 19.7 & 12.3 & 36.8 & 26.3 & 34.9 \\
Tongyi DeepResearch-30B-A3B & Qwen2.5-30B & 46.5 & 89.5 & \textbf{96.4} & -- & -- & 66.7 & 40.6 & 39.1 & 34.3 & 46.8 & \textbf{45.4} & 51.3 \\
\midrule
\multicolumn{14}{@{}l}{\cellcolor{headerRow}\textbf{\textit{Fixed Pipeline Deep Research}}} \\
WebThinker QwQ-32B (report) & QwQ-32B & 45.2 & 86.4 & 94.3 & -- & -- & 72.8 & 37.9 & 36.2 & 32.6 & 43.2 & 42.9 & 52.0 \\
WebThinker-32B-DPO (report) & Qwen2.5-32B & 46.7 & 91.2 & 95.5 & -- & -- & 74.2 & 40.6 & 39.4 & 35.4 & 46.0 & 43.5 & 53.8 \\
\rowcolor{closedRow} \aitwo\ Ai2 ScholarQA & Claude Sonnet & 87.7 & 88.1 & 89.1 & 92.4 & 81.2 & 75.0$^\dagger$ & 36.1 & 35.1 & 32.0 & 40.5 & 38.9 & 66.3 \\
\midrule
\multicolumn{14}{@{}l}{\cellcolor{headerRow}\textbf{\textit{Open Deep Research}}} \\
Qwen3-8B + Search & Qwen3-8B & 57.2 &  42.8 & 92.1 & 53.7 & 40.3 & 46.3 & 18.2 &  14.3 & 8.7 & 29.5 &  24.4 & 40.6 \\
\aitwo\ DR Tulu-8B (SFT) \openmodel\ \opencode & Qwen3-8B & 72.3 & 81.4 & 91.0 & 65.3 & 51.6 & 68.5 & 39.0 & 36.3 & 35.3 & 45.5 & 39.5 & 59.9 \\
\aitwo\ DR Tulu-8B (1000-step, $\approx$6000h) \openmodel\ \opencode & Qwen3-8B & 86.7 & 84.8 & 95.4 & 90.6 & \textbf{76.1} & 71.1 & 41.8 & 39.5 & 39.4 & 47.3 & 41.6 & 66.5 \\
\aitwo\ DR Tulu-8B (1900-step, $\approx$9700h) \openmodel\ \opencode & Qwen3-8B & \textbf{86.8} & \textbf{89.6} & 95.4 & 88.6 & \textbf{73.7} & 74.3 & 43.4 & \textbf{41.7} & \textbf{41.8} & 48.2 & 41.3 & 68.2 \\
\rowcolor{oursRow}
\evtree\ \textbf{Ours (SFT)} \openmodel\ \opencode & Qwen3-8B & 79.5 & 78.4 & 94.2 & 83.9 & 61.4 & 64.8 & 38.0 & 33.8 & 36.0 & 43.5 & 40.1 & 60.8 \\
\rowcolor{oursRow}
\evtree\ \textbf{Ours (75-step, $\approx$402h)} \openmodel\ \opencode & Qwen3-8B & 85.1 & 85.3 & 93.7 & 89.3 & 72.1 & 74.2 & 41.9 & 39.7 & 40.0 & 46.9 & 41.5 & 67.1 \\
\rowcolor{oursRow}
\evtree\ \textbf{Ours (140-step, $\approx$750h)} \openmodel\ \opencode & Qwen3-8B & 86.0 & 86.4 & 94.0 & \textbf{91.5} & 72.2 & \textbf{75.2} & \textbf{43.6} & \textbf{41.7} & \textbf{41.8} & \textbf{48.4} & 42.5 & \textbf{68.3} \\
\bottomrule
\end{tabular}%
}

\caption{\textbf{Main results with expanded SQAv2 and DRB columns.} The average is computed over the displayed overall scores (SQAv2, ResearchQA, and DRB) when all three are available. Rows with \colorbox{closedRow}{gray background} use proprietary models. \colorbox{oursRow}{Green} highlights our method. \openmodel\ indicates open model weights and training data; \opencode\ indicates open training code. \textbf{Bold numeric values} mark the best score among open models for that column. $^*$Scores reported by original benchmark authors. $^\dagger$Evaluated on 100-sample subset. ``--'' = not reported.}
\label{tab:main_results}
\end{table*}

\noindent\textbf{Baselines.}
We organize baselines into three groups in Table~\ref{tab:main_results}: (1)~non-rubric open systems, including Qwen3-8B/QwQ-32B RAG~\citep{qwen3,qwq}, Search-R1~\citep{searchr1}, ASearcher~\citep{asearcher}, WebExplorer~\citep{webexplorer}, WebThinker and its report-mode variants~\citep{webthinker}, Tongyi DeepResearch~\citep{tongyi_dr}, and Ai2 ScholarQA~\citep{singh-etal-2025-ai2}; (2)~matched rubric-RL systems, especially DR~Tulu-8B~\citep{drtulu_rler}, which shares our Qwen3-8B scale and GRPO-style training but uses query-conditioned rubrics; and (3)~Proprietary reference systems, including Perplexity, Gemini, GPT-5~\citep{gpt5}, and OpenAI Deep Research.
Open baselines use official code and recommended configurations; additional setup details are in Appendix~\ref{app:baselines}.

\noindent\textbf{Implementation Details.}

\textbf{Data construction.}
We build evidence trees from two source corpora: Wikipedia \footnote{\url{https://huggingface.co/datasets/inclusionAI/ASearcher-Local-Knowledge}} and OpenScholar~\citep{openscholar}\footnote{\url{https://huggingface.co/datasets/OpenSciLM/OpenScholar-DataStore-V3}}.
To control annotation cost, we use DeepSeek-V3.2~\citep{deepseek_v3} for the high-volume stages, including evidence-tree expansion from seed topics and query--rubric synthesis from each tree.
This yields approximately 10K candidate pairs.
We reserve GPT-5.1~\citep{gpt5} for verification, which checks evidence support, query--rubric alignment, and criterion redundancy.
After revision, 9{,}064 pairs are retained and used for SFT and RL training.

\textbf{Training.}
We initialize from Qwen3-8B~\citep{qwen3}.
To adapt the model to the expected tool-use and report format, we perform a lightweight cold-start SFT stage on our 200 generated queries.
GPT-5.1 annotates reference trajectories for these queries, including reasoning steps, search actions, and final report writing, and we fine-tune for 2 epochs (${\sim}$3 GPU-hours).
We then train the SFT checkpoint with GRPO on the remaining 8{,}864 query--rubric pairs for 140 steps, corresponding to approximately one epoch (${\sim}$750 GPU-hours), using Qwen3.5-35B-A3B as the rubric reward judge.
All training uses 8$\times$A100 80GB GPUs; hyperparameters are reported in Appendix~\ref{app:hyperparams}.

\textbf{Tool environment.}
During training, the agent uses three local tools backed by the same Wikipedia and OpenScholar corpora used for data construction: \texttt{search} performs dense retrieval over the local Wikipedia corpus, \texttt{browse} fetches full article text from the local Wikipedia database, and \texttt{scholar} performs dense retrieval over OpenScholar papers.
This local environment avoids external API calls during RL rollouts, reducing latency, eliminating per-step search costs, and improving run-to-run stability.
For evaluation, we follow prior work~\citep{drtulu_rler} and use the same live tools for fair comparison: \texttt{search} calls Google Search via the Serper API, \texttt{browse} reads web pages through Jina Reader, and \texttt{scholar} queries Google Scholar via Serper API.

\subsection{Main Results}
\label{sec:main_results}


Table~\ref{tab:main_results} reports performance across the three benchmarks, together with the intermediate checkpoints of our own run.
The SFT cold start reaches a 60.8 average, 75 RL steps (${\approx}$402 GPU-hours) raise this to 67.1, and 140 steps (${\approx}$750 GPU-hours) to 68.3.
Most of the gain therefore arrives in the first 75 steps, and the remaining 1.2 points cost roughly as much compute again---which is why we treat 140 steps as the stopping point and analyze the step budget separately in \S\ref{sec:training_efficiency}.

\paragraph{Comparison with open models.}

\textsc{DeepRubric}-8B achieves the strongest average among open deep research systems in our evaluation.
Relative to the Qwen3-8B search-only base agent, it raises the three-benchmark average from 40.6 to 68.3, with large gains on SQAv2 (57.2$\to$86.0), ResearchQA (46.3$\to$75.2), and DRB (18.2$\to$43.6).
The closest comparison is DR~Tulu-8B~(RL), which uses the same Qwen3-8B backbone family, GRPO-style training, and rubric-reward interface, but evolves search-grounded rubrics from policy rollouts.
In contrast, \textsc{DeepRubric} uses the fixed rubrics generated and verified before RL, yet reaches a similar average (68.3 vs.\ 68.2) with 140 RL steps and 750 GPU-hours, compared with DR~Tulu's 1{,}900-step, 9{,}700 GPU-hour run.
\textsc{DeepRubric} is higher on ResearchQA (75.2 vs.\ 74.3) and DRB (43.6 vs.\ 43.4), but slightly lower on SQAv2 (86.0 vs.\ 86.8), where DR~Tulu has higher rubric coverage and citation recall.
It attains the best open-model scores on ResearchQA, DRB overall, and DRB instruction following, while also improving SQAv2 citation precision over DR~Tulu (91.5 vs.\ 88.6).

\paragraph{Comparison with Proprietary systems.}

\textsc{DeepRubric}-8B is competitive with proprietary systems on scientific QA, while the strongest proprietary systems remain ahead on broader report generation.
On SQAv2, \textsc{DeepRubric} exceeds the proprietary deep-research references in the upper block, including GPT-5~+~Search and OpenAI Deep Research; the only higher score in the table comes from Ai2 ScholarQA, a fixed pipeline built on Claude Sonnet.
On ResearchQA and DRB, the best proprietary systems still lead, especially on DRB report-quality dimensions such as comprehensiveness, depth, and readability.
This pattern suggests that evidence-tree supervision can make an 8B open model highly competitive on rubric- and citation-oriented scientific QA, while open-ended report generation still benefits from stronger base models and retrieval infrastructure.
The expanded SQAv2 and DRB columns help localize these effects: rubric coverage, comprehensiveness, and depth are closest to our supervision signal, whereas citation precision/recall, instruction-following, and readability also depend on auxiliary rewards and inference-time formatting choices.

\subsection{Training Efficiency Analysis}
\label{sec:training_efficiency}


\begin{table}[t]
\centering
\small
\resizebox{\columnwidth}{!}{
\renewcommand{\arraystretch}{1.2}
\setlength{\tabcolsep}{4pt}
		\begin{tabular}{l c c c | c}
		\toprule
		     \textbf{Variant} & SQAv2 & ResearchQA & DRB & Avg. \\
	\midrule
	    \rowcolor{oursRow}
	    \textbf{Ours (full)}
	    & 85.1
	    & 74.2
	    & 41.9
	    & 67.1 \\
	    \quad w/ DeepSeek-V3.2 revision
	    & 84.0
	    & 76.0
	    & 41.1
	    & 67.0$_{-0.1\%}$ \\
	    \quad w/o revision
	    & 83.8
	    & 73.2
	    & 39.9
	    & 65.6$_{-2.2\%}$ \\
	    \quad Search-based rubrics
	    & 80.6
	    & 72.1
	    & 37.3
	    & 63.3$_{-5.7\%}$ \\
	    \quad Closed-book rubrics
	    & 83.2
	    & 70.4
	    & 40.7
	    & 64.8$_{-3.4\%}$ \\
	\bottomrule
	\end{tabular}
	}
				\caption{\textbf{Ablation on rubric construction and verification (75 steps).} All variants share the same training queries, local retrieval environment, reward composition, and 75-step GRPO recipe, differing only in how rubrics are constructed or revised.
                }
	\label{tab:ablation}
	\end{table}

\paragraph{Step efficiency.}

\begin{figure}[t]
\centering
\includegraphics[width=\columnwidth]{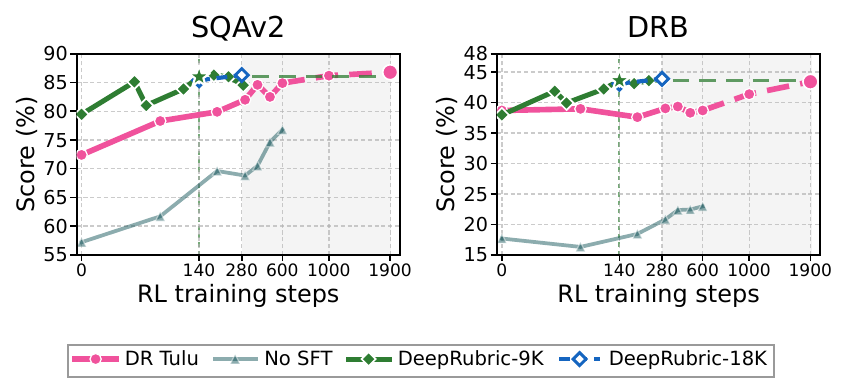}
\caption{\textbf{Training-step efficiency.} \textsc{DeepRubric} reaches competitive performance with fewer RL steps.}
\label{fig:step_efficiency}
\end{figure}

Figure~\ref{fig:step_efficiency} compares checkpoint performance over the course of RL training for three configurations: \textsc{DeepRubric}-9K (green, ours), DR~Tulu (pink, full RLER with an SFT cold start), and No SFT (gray, DR~Tulu-style RL initialized directly from the base model). \textsc{DeepRubric}-9K improves rapidly, matching the three-benchmark average achieved by DR~Tulu at its 1,900-step endpoint after only 140 steps (68.3 vs.\ 68.2). This corresponds to \(13.6\times\) fewer policy updates and \(6.8\times\) fewer rollouts.
Further training on the 9K pool provides no additional benefit and can even degrade performance once training extends beyond approximately one epoch. We therefore expand the training pool to 18K examples to test whether a longer RL schedule remains effective with more data. On the expanded pool, the 280-step checkpoint reaches 86.3 on SQAv2 and 43.9 on DRB, yielding a three-benchmark average of 69.4 and showing that further RL training can continue to improve performance when more data are available. Detailed checkpoint-level and data-scaling results are provided in Appendix~\ref{app:data_scaling}.

\paragraph{Training cost.}
Table~\ref{tab:training_cost} summarizes the end-to-end cost.
Our total training budget is ${\sim}$\$1.7K (including \$180 in API calls, 3 GPU-hours for SFT, and 750 GPU-hours for RL on 8$\times$A100), compared to ${\geq}$\$30K for DR~Tulu---a ${\sim}$17$\times$ reduction. GPU costs are estimated using rental server pricing.
The reduction is driven by higher reward signal quality per rollout: evidence-tree rubrics provide denser, more structured feedback, enabling the policy to converge in 140 steps rather than 1{,}900, which directly translates to ${\sim}$13$\times$ fewer RL GPU-hours. Because GPU-hours also depend on retrieval latency and accelerator type, Appendix~\ref{app:efficiency_analysis} provides a detailed decomposition and reports latency- and hardware-independent measures: \(13.6\times\) fewer policy updates and \(6.8\times\) fewer rollouts.

\begin{figure*}[t]
\centering
\includegraphics[width=\textwidth]{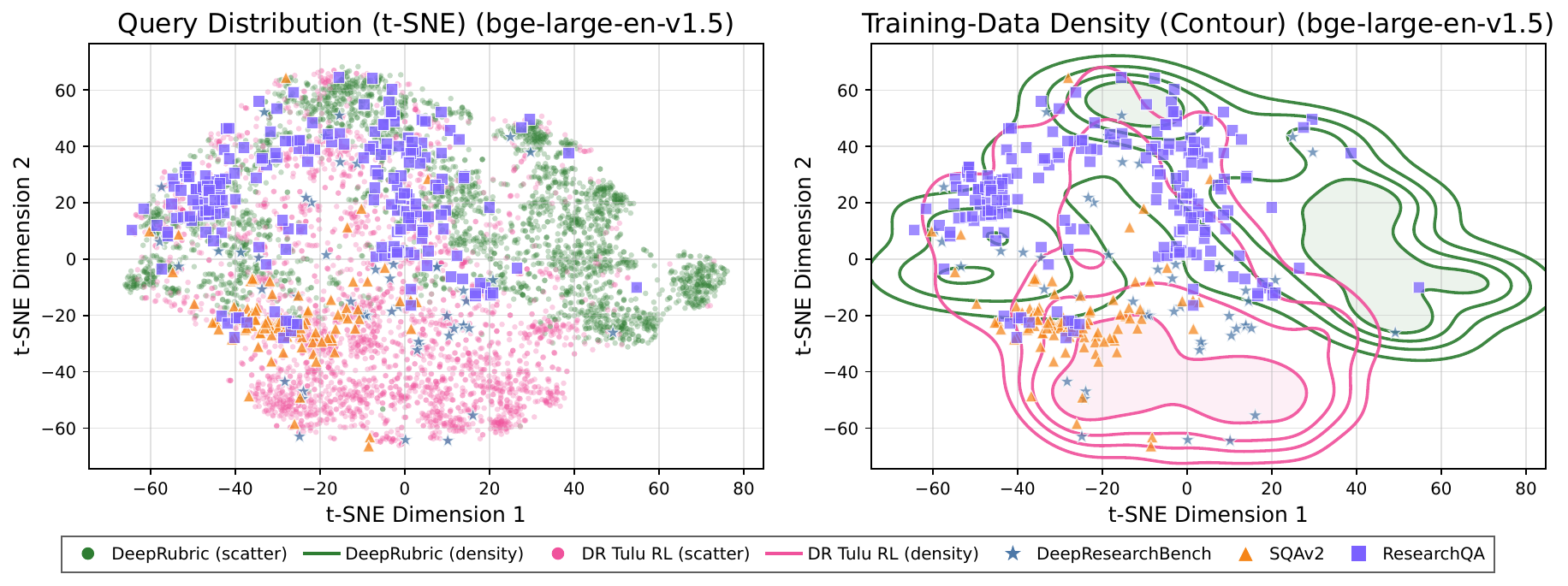}
\caption{\textbf{Training query distribution.} BGE-large embeddings projected with t-SNE. Left: scatter plot of training queries and benchmark queries. Right: density contours of training data with benchmark queries overlaid. 
}
\label{fig:query_semantic_coverage}
\end{figure*}

\begin{table}[t]
\centering
\scriptsize
\renewcommand{\arraystretch}{1.15}
\setlength{\tabcolsep}{2pt}
\resizebox{\columnwidth}{!}{%
\begin{tabular}{@{}lccccr@{}}
\toprule
\textbf{Method} & \textbf{Data annotation} & \textbf{SFT} & \textbf{Rollouts} & \textbf{RL training} & \textbf{Est.\ cost} \\
\midrule
\aitwo\ DR Tulu-8B & GPT-5 (16K traj.)$^\dagger$ & 136 GPU-hrs & 486{,}400 & 9{,}700 GPU-hrs & $\geq$\$30K \\
\rowcolor{oursRow}
\evtree\ \textbf{Ours (8B)} & \$180 (API) & 3 GPU-hrs & \textbf{71{,}680} & \textbf{750 GPU-hrs} & $\sim$\$1.7K \\
\bottomrule
\end{tabular}
}
\caption{\textbf{Training cost comparison.} Rollouts are a latency- and hardware-independent measure of convergence efficiency.
}
\label{tab:training_cost}
\end{table}

\subsection{Ablation Study}
\label{sec:ablation}

To isolate the effect of evidence-tree construction, we conduct a fixed-query counterfactual ablation: the training queries are kept unchanged, while the rubrics are regenerated without access to the original evidence tree. This setup tests whether the tree structure provides supervision beyond what can be recovered from the final query alone. We compare three variants: \textbf{Closed-book rubrics}, which generate criteria directly from the query; \textbf{Search-based rubrics}, which first retrieve passages using the query and then generate criteria from the retrieved context; and \textbf{w/o revision}, which keeps the tree-based rubrics but removes the LLM-based verification pass. For cost efficiency, all variants are trained for 75 GRPO steps under the same recipe (Table~\ref{tab:ablation}). The full pipeline outperforms both query-first rubric baselines by 2.3--3.8 points on average, showing that adding retrieved passages to query-first rubric generation does not recover the benefit of tree-structured decomposition.

\paragraph{How much of the supervision comes from the verifier?}
Since 91.5\% of retained pairs are revised by GPT-5.1, one may ask whether the supervision mainly reflects a strong proprietary model rather than the evidence tree; replacing it with the open-weights DeepSeek-V3.2 yields 67.0 against 67.1, and removing revision entirely still leaves tree-based rubrics ahead of both query-first variants (65.6 vs.\ 64.8 and 63.3), so the tree carries most of the signal (Appendix~\ref{app:verify_stats}).

\subsection{Direct Evaluation of Rubric Quality}
\label{sec:pairacc}

\begin{table}[t]
\centering
\small
\resizebox{\columnwidth}{!}{
\renewcommand{\arraystretch}{1.2}
\setlength{\tabcolsep}{4pt}
\begin{tabular}{l c c c}
\toprule
\textbf{Rubric source} & \textbf{PairAcc} & \multicolumn{2}{c}{\textbf{Undetected corruptions}} \\
 & (\%) & (\%) & (of 750) \\
\midrule
\rowcolor{oursRow}
Evidence tree (ours) & \textbf{93.87} & \textbf{6.13} & \textbf{46} \\
Search-based (query-first) & 86.80 & 13.20 & 99 \\
\midrule
$\Delta$ & $+$7.07 & $-$7.07 & $-$53 \\
\bottomrule
\end{tabular}
}
\caption{\textbf{Rubric quality measured directly on the data.} PairAcc measures how often the original report scores above its corrupted counterpart across 750 pairs.}
\label{tab:pairacc}
\end{table}

Because end-task performance is only an indirect proxy for supervision quality, we directly test whether a rubric scores a factual report above a corrupted version of the same report. For 750 queries, we generate reports with \textsc{DeepRubric}-8B and construct corrupted counterparts by replacing 3--5 factual claims, such as names, dates, years, and negations, with plausible but incorrect alternatives. Each pair is scored with both the evidence-tree and query-first rubrics synthesized for the same query, using an identical judge and scoring protocol (Appendix~\ref{app:pairacc}).
As shown in Table~\ref{tab:pairacc}, Evidence-tree rubrics achieve a PairAcc of 93.87\%, compared with 86.80\% for query-first rubrics, a gain of 7.07 points (paired-bootstrap 95\% CI: [5.20, 8.93]). They more than halve undetected corruptions, from 99 of 750 pairs (13.20\%) to 46 (6.13\%). This result shows that criteria anchored to specific evidence provide a more discriminative factual reward signal than criteria inferred from the query alone.

\subsection{Training Data Analysis}
\label{sec:generated_data_analysis}

\paragraph{Construction statistics.}
Starting from 9{,}838 evidence trees (depth 3, ${\sim}$55 nodes each), the LLM-based verification produces 9{,}064 retained query--rubric pairs (92.1\% retention rate): among these, 91.5\% are revised for improved query--rubric alignment and 0.6\% are kept unchanged; the remaining 7.9\% are dropped.
Each retained pair contains on average 7.0 rubric criteria (58.4\% factual, 41.6\% logical) grounded in 5.0 selected evidence leaves with 2.9 passages per leaf.
Full statistics are reported in Appendix~\ref{app:data_stats}.

\paragraph{Semantic Distribution.}

Figure~\ref{fig:query_semantic_coverage} projects BGE-large query embeddings for \textsc{DeepRubric}, DR~Tulu RL, and three evaluation benchmarks via t-SNE.
The two training sets occupy complementary regions despite targeting the same task, and their density shapes differ: \textsc{DeepRubric} covers a wider area with lower peak density, reflecting greater query diversity from multi-leaf evidence-tree synthesis; DR~Tulu RL concentrates in a narrower region, reflecting more homogeneous single-document queries.
The benchmark overlays connect this to downstream performance: SQAv2 queries fall within the DR~Tulu RL density region (both methods perform comparably there), while ResearchQA and DRB queries have greater presence near the \textsc{DeepRubric}-dense region, consistent with our gains on these benchmarks.
This suggests the advantage of evidence-tree co-generation is \emph{structural}---it produces more diverse training queries that teach the model to decompose complex information needs---rather than merely topical.

\subsection{Case Study}
\label{sec:case_study}

Table~\ref{tab:case_study_final_response} presents a representative DRB case, including the prompt, abridged final answers, and DRB scores.
The prompt requires a multi-faceted analysis of investments, products, client cases, strategy, talent programs, and future trends, rather than a flat enumeration of firms.
DR~Tulu-8B mainly organizes its response by first identifying a set of firms and then accumulating firm-specific evidence, yielding a broad inventory but making the cross-firm implications less explicit.
In contrast, \textsc{DeepRubric}-8B begins with a higher-level pattern and concludes by assigning firms to comparative roles, such as platform builders, large-scale adopters, and strategy/governance specialists.

\section{Conclusion}
\label{sec:conclusion}


We introduced \textsc{DeepRubric}, an evidence-first framework for constructing query--rubric supervision for reinforcement learning of deep research agents. By deriving both training queries and reward criteria from the same evidence tree, \textsc{DeepRubric} grounds rubrics in retrieved documents and aligns rewards with the intended task scope. Experiments on three open-retrieval benchmarks show that an 8B model trained with \textsc{DeepRubric} approaches the strongest open rubric-RL baseline with substantially fewer RL steps, rollouts, and GPU-hours, while ablations, a direct evaluation of the synthesized rubrics, and a supervision-scaling study further validate the effectiveness of evidence-tree construction for reward supervision.

\section*{Limitations}
\label{sec:limitations}

\textsc{DeepRubric} is limited by the corpora used to construct supervision.
Our current data is built from Wikipedia and OpenScholar, which provide broad factual and scholarly coverage but may not fully support specialized domains such as clinical, legal, or proprietary enterprise research.
Moreover, because training queries are synthesized from corpus evidence rather than sampled from real user demand, the resulting distribution is corpus-shaped and may not cover all deployed deep-research scenarios equally well, especially those involving private data, domain-specific standards, or rapidly changing information.

\section*{Acknowledgements}

This work is supported by the Zhongguancun Academy (Grant No.\ C20250210), the National Natural Science Foundation of China (Grant Nos.\ 62372275, 62472261, 62502282), the Natural Science Foundation of Jinan (Grant Nos.\ JNSX2025004, 202534077), and the Key R\&D Program of Shandong Province (Grant No.\ 2025CXGC010112).


\bibliography{references}

\clearpage
\newpage
\appendix

\section*{Appendix}

\addcontentsline{toc}{section}{Appendix}

\section{Reward Function Details}
\label{app:reward_details}

We provide additional details of each reward component in the composite reward (Eq.~\ref{eq:total_reward}).

\subsection{Rubric Scoring}
\label{app:rubric_scoring}

For each leaf criterion in the evidence tree, we use an LLM judge (Qwen3.5-35B-A3B) to evaluate whether the generated report adequately addresses the criterion.
The judge receives the original query, the model's response, and a single criterion (including criterion ID, type, description, and supporting evidence for factual criteria), and returns a score on an integer scale from 0 to 4.
The score is normalized to $[0, 1]$ by dividing by 4.
We adopt a structured rubric scoring prompt adapted from DR~Tulu~\citep{drtulu_rler}, shown in Figure~\ref{fig:rubric_judge_prompt}.

\begin{center}
\begin{promptbox}[title=Rubric Scoring Judge Prompt]
\footnotesize\ttfamily
\textbf{[System]}\\[2pt]
You will be given a question (in \texttt{<question>} tags), an answer (in \texttt{<response>} tags), and a single criterion (in \texttt{<criterion>} tags). Judge only how well the answer satisfies that specific criterion.\\[4pt]
The criterion can be of different types:\\
\hangindent=1em -- \textbf{Factual}: judge whether the answer provides the specific, correct, and relevant factual content required. Do not reward merely mentioning related topics.\\
\hangindent=1em -- \textbf{Logical}: judge whether the answer performs the specific reasoning required---comparison, distinction, synthesis, or conclusion.\\[4pt]
Return ONLY a JSON object \texttt{\{"score": x\}} where x is an integer from 0 to 4.\\[4pt]
\textbf{Scoring guidelines:}\\
\hangindent=1em 4: Fully satisfies. All essential components explicitly, correctly addressed.\\
\hangindent=1em 3: Mostly satisfies. Minor omissions or ambiguities.\\
\hangindent=1em 2: Partially satisfies. One or more essential parts missing.\\
\hangindent=1em 1: Minimally satisfies. Mostly generic, weak, or poorly connected.\\
\hangindent=1em 0: Does not satisfy. Required content missing or incorrect.\\[6pt]
\textbf{[User]}\\[2pt]
\texttt{<question>\{question\}</question>}\\
\texttt{<response>\{answer\}</response>}\\
\texttt{<criterion>\{criterion\_text\}</criterion>}
\end{promptbox}
\captionof{figure}{\textbf{Rubric scoring judge prompt.} For factual criteria, the criterion text includes the criterion type, description, and supporting evidence from the evidence tree, enabling the judge to verify factual claims against specific sources.}
\label{fig:rubric_judge_prompt}
\end{center}

The per-criterion scores are aggregated into a single rubric reward via a weighted average over all criteria:
\begin{equation}
R_{\mathrm{rubric}}(y) = \frac{\sum_{r \in \mathcal{R}(T)} w_r \cdot s_r(y)}{\sum_{r \in \mathcal{R}(T)} w_r},
\end{equation}
where $s_r(y) \in [0, 1]$ is the normalized judge score for criterion $r$ and $w_r$ is its importance weight.

\subsection{Format Reward}
\label{app:format_reward}

The format reward $R_{\mathrm{format}}(y) \in [0, 1]$ checks whether the model output follows the expected structured format.
It is computed as a weighted sum of four binary indicators:
\begingroup\small
\begin{equation}
\begin{split}
R_{\mathrm{format}}(y) ={}& 0.5 I_{\mathrm{answer}}(y)
 + 0.2 I_{\mathrm{cite}}(y) \\
& + 0.1 I_{\mathrm{tool}}(y)
 + 0.2 I_{\mathrm{think}}(y),
\end{split}
\end{equation}
\endgroup
where $I_{\mathrm{answer}}(y)$ indicates the presence of properly formed \texttt{<answer>...</answer>} tags containing the final report, $I_{\mathrm{cite}}(y)$ indicates the presence of inline citations, $I_{\mathrm{tool}}(y)$ indicates at least one valid tool call was issued, and $I_{\mathrm{think}}(y)$ indicates the model produced reasoning blocks.
The largest weight is assigned to the answer tag, as it is the most critical structural requirement.

\subsection{Search Turn Reward}
\label{app:search_reward}

The search turn reward $R_{\mathrm{search}}(y) \in [0, 1]$ encourages the agent to issue a sufficient number of search queries during report generation:
\begin{equation}
R_{\mathrm{search}}(y) = \min\!\left(\frac{N_{\mathrm{search}}(y)}{U}, 1.0\right),
\end{equation}
where $N_{\mathrm{search}}(y)$ is the number of valid tool calls (across \texttt{search}, \texttt{browse}, and \texttt{scholar} tools) extracted from \texttt{<tool\_call>} JSON blocks, and $U = 6$ is the upper bound.
The reward saturates at $U$ calls to avoid rewarding excessive searching.

\subsection{Citation Reward}
\label{app:citation_reward}

The citation reward $R_{\mathrm{cite}}(y) \in [0, 1]$ evaluates the quality of inline citations in the generated report, adapted from DR~Tulu~\citep{drtulu_rler}.
Unlike simple citation-format checks, we employ a per-claim evaluation framework that assesses both the \emph{content quality} of citations (do they support the claims?) and the \emph{structural quality} (are they well-distributed and properly formatted?).

\subsubsection{Claim Extraction}
\label{app:claim_extraction}
We first segment the generated answer into individual claims.
Each \texttt{<cite id="ID">text</cite>} tag defines a cited claim paired with its citation IDs; text outside cite tags is treated as an uncited claim.
This yields a set of claims $\mathcal{C} = \{(c_i, \mathcal{S}_i)\}_{i=1}^{|\mathcal{C}|}$, where $c_i$ is the claim text and $\mathcal{S}_i \subseteq \mathcal{D}$ is the set of cited source IDs (empty for uncited claims), and $\mathcal{D}$ is the citation store of all retrieved snippets.

\subsubsection{Per-Claim F$_1$ Score}
\label{app:per_claim_f1}
For each claim $c_i$ we compute recall and precision scores using an LLM judge (same Qwen3.5-35B-A3B used for rubric scoring):

\paragraph{Citation recall.}
For cited claims ($\mathcal{S}_i \neq \varnothing$), we concatenate the source texts and ask the judge to assess support on a three-point scale:
\begin{equation}
\text{recall}(c_i) = \begin{cases}
1.0 & \text{if \textit{Fully supported}} \\
0.5 & \text{if \textit{Partially supported}} \\
0.0 & \text{if \textit{No support}}
\end{cases}
\end{equation}
For uncited claims ($\mathcal{S}_i = \varnothing$), we instead ask whether the claim \emph{requires} a citation.
If the judge determines the claim is a factual statement that needs grounding, the recall score is 0 (penalizing the missing citation); otherwise it is 1 (the claim is a transition, summary, or reasoning step that does not need citation).

\paragraph{Citation precision.}
For cited claims, the judge assesses whether the cited snippets contain key information relevant to the claim:
\begin{equation}
\text{precision}(c_i) = \begin{cases}
1.0 & \text{if \textit{Relevant}} \\
0.0 & \text{if \textit{Irrelevant}}
\end{cases}
\end{equation}
Uncited claims receive a precision of 1 (no irrelevant citation was introduced).

\paragraph{Per-claim F$_1$.}
The per-claim score combines recall and precision via the harmonic mean:
\begin{equation}
F_1(c_i) = \frac{2 \cdot \text{recall}(c_i) \cdot \text{precision}(c_i)}{\text{recall}(c_i) + \text{precision}(c_i)}.
\end{equation}
The average per-claim F$_1$ across all claims is:
\begin{equation}
\bar{F}_1 = \frac{1}{|\mathcal{C}|} \sum_{i=1}^{|\mathcal{C}|} F_1(c_i).
\end{equation}

\subsubsection{Citation Format Reward}
\label{app:citation_format}
Beyond content quality, we reward well-structured citation usage via the citation format reward $R_{\mathrm{fmt}}$.
This score captures three aspects:

\paragraph{Citation ID validity.}
Let $\mathcal{A}$ be the set of all citation IDs referenced in the answer.
The ID validity ratio is $v = |\{a \in \mathcal{A} : a \in \mathcal{D}\}| / |\mathcal{A}|$, measuring the fraction of cited IDs that resolve to actual retrieved snippets.

\paragraph{Meaningful claim ratio.}
Among cited claims, we filter out trivially short or symbol-dominated spans.
The meaningful claim ratio $m$ is the fraction of cited claim texts that pass lexical quality checks (at least 4 lexical units and 18 characters).
The base quality is $Q = v \cdot m$.

\paragraph{Citation distribution.}
We compute the normalized position of each unique valid citation within the answer text and score the distribution on three sub-metrics:
(i)~\emph{span}: the range covered by citations (normalized by 0.6);
(ii)~\emph{uniformity}: deviation from ideal even spacing; and
(iii)~\emph{center}: proximity of mean citation position to 0.5.
The distribution score is $d = 0.4 \cdot \text{span} + 0.4 \cdot \text{uniformity} + 0.2 \cdot \text{center}$.

\paragraph{Citation count.}
The count score rewards using a sufficient number of distinct valid citations: $n = \min(|\{a \in \mathcal{A} : a \in \mathcal{D}\}| / 6,\, 1.0)$.

The final citation format reward is:
\begin{equation}
R_{\mathrm{fmt}} = Q \cdot (0.7 + 0.1 \cdot d + 0.2 \cdot n).
\end{equation}

\subsubsection{Final Citation Reward}
\label{app:final_citation}
The overall citation reward combines content quality and structural quality:
\begin{equation}
R_{\mathrm{cite}}(y) = 0.6 \cdot \bar{F}_1 + 0.4 \cdot R_{\mathrm{fmt}}.
\end{equation}

The citation judge prompts are shown in Figures~\ref{fig:citation_recall_prompt}--\ref{fig:citation_precision_prompt}.

\begin{center}
\begin{promptbox}[title=Citation Recall Prompt (Cited Claims)]
\footnotesize\ttfamily
You are an expert in evaluating text quality. You will receive a user's question, a factual statement from an AI assistant's response, and a snippet from the document. Assess whether this statement is supported by the snippet.\\[4pt]
\hangindent=1em -- \textbf{[[Fully supported]]}: Most information in the statement is supported by or extracted from the snippet. Statement and snippet parts are almost identical.\\
\hangindent=1em -- \textbf{[[Partially supported]]}: More than half of the content is supported, but a small portion is not mentioned or contradicts the snippet.\\
\hangindent=1em -- \textbf{[[No support]]}: The statement is largely unrelated, or most key points do not align.\\[4pt]
Provide: ``Rating: [[...]] Analysis: ...''\\[4pt]
\texttt{<question>\{question\}</question>}\\
\texttt{<statement>\{statement\}</statement>}\\
\texttt{<snippet>\{cited\_snippets\}</snippet>}
\end{promptbox}
\captionof{figure}{\textbf{Citation recall judge prompt} for cited claims with source snippets.}
\label{fig:citation_recall_prompt}
\end{center}

\begin{center}
\begin{promptbox}[title=Citation Recall Prompt (Uncited Claims)]
\footnotesize\ttfamily
You are an expert in evaluating citation coverage for source-grounded QA.\\[4pt]
You will receive: (1) the user's question, (2) the assistant's final answer, (3) one sentence from the final answer.\\[4pt]
Determine whether the sentence is a factual claim that should be grounded with an explicit citation.\\[4pt]
Answer \textbf{[[Yes]]} if the sentence introduces a concrete factual claim, empirical result, dataset/model detail, number, comparison, or attribution.\\
Answer \textbf{[[No]]} if the sentence is only an introductory phrase, transition, restatement of the question, or summary following already cited claims.\\[4pt]
Output: ``Need Citation: [[Yes/No]] Analysis: ...''\\[4pt]
\texttt{<question>\{question\}</question>}\\
\texttt{<final\_answer>\{full\_response\}</final\_answer>}\\
\texttt{<statement>\{statement\}</statement>}
\end{promptbox}
\captionof{figure}{\textbf{Citation recall judge prompt} for uncited claims. Claims judged as needing citation but lacking one receive a recall score of 0.}
\label{fig:citation_nocite_prompt}
\end{center}

\begin{center}
\begin{promptbox}[title=Citation Precision Prompt]
\footnotesize\ttfamily
You are an expert in evaluating text quality. Assess whether the snippet contains key information of the statement.\\[4pt]
\hangindent=1em -- \textbf{[[Relevant]]}: Some key points of the statement are supported by or extracted from the snippet.\\
\hangindent=1em -- \textbf{[[Irrelevant]]}: The statement is almost unrelated to the snippet.\\[4pt]
Provide: ``Rating: [[...]] Analysis: ...''\\[4pt]
\texttt{<question>\{question\}</question>}\\
\texttt{<statement>\{statement\}</statement>}\\
\texttt{<snippet>\{cited\_snippets\}</snippet>}
\end{promptbox}
\captionof{figure}{\textbf{Citation precision judge prompt.} Evaluates whether each cited snippet is relevant to the claim it supports.}
\label{fig:citation_precision_prompt}
\end{center}

\section{Training Details}
\label{app:hyperparams}

Our training infrastructure is built on \texttt{verl-tool}~\citep{verltool}, an open-source framework for training tool-augmented language models with reinforcement learning.
The reward scoring pipeline is adapted from DR~Tulu~\citep{drtulu_rler}.
All training runs use 8$\times$A100 80GB GPUs on a single node.

\subsection{SFT Hyperparameters}
\label{app:sft_hyperparams}

We provide the hyperparameters used during the SFT cold-start stage in Table~\ref{tab:sft_hypers}.

\begin{table}[t]
\small
\centering
\resizebox{\columnwidth}{!}{%
\begin{tabular}{lc}
\toprule
\textbf{Hyperparameter} & \textbf{Value} \\
\midrule
Base model & Qwen3-8B \\
SFT data size & 200 examples \\
Annotation model & GPT-5.1 \\
Number of training epochs & 2 \\
Learning rate & $4 \times 10^{-5}$ \\
Learning rate scheduler & cosine \\
Warmup ratio & 0.1 \\
Per-device batch size & 1 \\
Gradient accumulation steps & 16 \\
Max sequence length & 16{,}384 \\
Data type & BF16 \\
Weight decay & 0.0 \\
\bottomrule
\end{tabular}
}
\caption{\textbf{Hyperparameters for SFT training.}}
\label{tab:sft_hypers}
\end{table}

\subsection{RL Hyperparameters}
\label{app:rl_hyperparams}

We provide the hyperparameters used during GRPO training in Table~\ref{tab:grpo_hypers}.

\begin{table}[t]
\small
\centering
\resizebox{\columnwidth}{!}{%
\begin{tabular}{lc}
\toprule
\textbf{Hyperparameter} & \textbf{Value} \\
\midrule
RL algorithm & GRPO \\
Base checkpoint & SFT checkpoint \\
Training data & 9{,}064 query-rubric pairs \\
Unique prompts per step & 64 \\
Rollouts per prompt (group size) & 8 \\
Rollouts per step & 512 \\
Mini-batch size & 4 \\
Max prompt length & 2{,}096 tokens \\
Max response length & 28{,}288 tokens \\
Max action length per tool call & 2{,}048 tokens \\
Max observation length per tool call & 4{,}096 tokens \\
Max tool call turns & 20 \\
Temperature & 1.0 \\
Top-$p$ & 1.0 \\
Learning rate & $1 \times 10^{-6}$ \\
Learning rate schedule & constant \\
KL penalty coefficient & 0.0 \\
Entropy coefficient & 0.0 \\
Total GRPO steps & 140 \\
Total rollouts & 71{,}680 \\
Checkpoint save frequency & every 5 steps \\
Data type & BF16 \\
Weight decay & 0.0 \\
Parallelism strategy & FSDP (param + optimizer offload) \\
vLLM GPU memory utilization & 0.6 \\
Rubric judge model & Qwen3.5-35B-A3B \\
\midrule
\multicolumn{2}{l}{\textit{Reward weights (Eq.~\ref{eq:total_reward})}} \\
\midrule
Rubric weight ($\alpha_{\text{rubric}}$) & 0.5 \\
Format weight ($\alpha_{\text{format}}$) & 0.2 \\
Citation weight ($\alpha_{\text{cite}}$) & 0.2 \\
Search weight ($\alpha_{\text{search}}$) & 0.1 \\
\bottomrule
\end{tabular}
}
\caption{\textbf{Hyperparameters for GRPO training.}  Training is performed on 8$\times$A100 80GB GPUs with FSDP and full parameter/optimizer offloading.  The reward weights follow the composite reward in Eq.~\ref{eq:total_reward}.}
\label{tab:grpo_hypers}
\end{table}

\section{Data Construction Details}
\label{app:data_construction}

We provide additional details of the evidence-tree data construction pipeline described in \S\ref{sec:tree}.

\subsection{Source Corpora}
\label{app:source_corpora}

We build evidence trees from two source corpora:
\begin{itemize}[nosep,leftmargin=*]
\item \textbf{Wikipedia (2018 dump):} We use the December 2018 Wikipedia dump as a general-knowledge retrieval corpus.  Seed topics are sampled from a diverse set of Wikipedia categories spanning science, history, technology, and social sciences.
\item \textbf{OpenScholar}~\citep{openscholar}: We use the OpenScholar corpus as a scientific-literature retrieval source.  Seed topics are sampled from academic research areas to ensure coverage of scientific deep research tasks.
\end{itemize}

Both corpora are indexed and exposed to the agent during data construction and RL training via three local retrieval tools: \texttt{search} (keyword search over the corpus), \texttt{browse} (full-text reading of a retrieved document), and \texttt{scholar} (semantic search over scientific papers).
During benchmark evaluation, the agent switches to the live search, browsing, and scholar tools described in \S\ref{sec:exp_setup}.

\subsection{Evidence Tree Generation Pipeline}
\label{app:tree_pipeline}

The evidence tree generation proceeds in the following steps:

\begin{figure*}[t]
\centering
\begin{minipage}{0.96\textwidth}
\small
\hrule
\vspace{0.35em}
\captionof{algorithm}{Evidence-Tree Query--Rubric Co-Generation}
\label{alg:pipeline}
\begin{algorithmic}[1]
\Require Corpus $\mathcal{C}$, retriever $\rho$, LLM $\mathcal{M}$, maximum depth $D_{\max}$
\Ensure Dataset $\mathcal{D} = \{(x_i, T_i, \mathcal{R}(T_i))\}_{i=1}^{N}$
\For{each sampled topic $t \sim \mathcal{C}$}
  \Statex \textit{Stage 1: top-down evidence-tree construction}
  \State Initialize root $v_0$ with $q_{v_0} \leftarrow t$ and $\mathcal{P}_{v_0} \leftarrow \rho(t)$
  \State $\textsc{Queue} \leftarrow \{v_0\}$
  \While{$\textsc{Queue} \neq \emptyset$}
    \State $v \leftarrow \textsc{Queue}.\mathrm{pop}()$
    \If{$\operatorname{depth}(v)=D_{\max}$ or $\mathcal{M}$ judges $q_v$ as atomic}
      \State \textbf{continue} \Comment{$v$ becomes a leaf}
    \EndIf
    \State $\mathcal{Q}_v \leftarrow \mathcal{M}(q_v, \mathcal{P}_v, \mathrm{Anc}(v))$
    \For{each child query $\tilde{q} \in \mathcal{Q}_v$}
      \State $\tilde{\mathcal{P}} \leftarrow \rho(\tilde{q})$
      \If{$\tilde{\mathcal{P}}$ is sufficient and $\tilde{q}$ is non-redundant}
        \State Create child $u$ with $q_u \leftarrow \tilde{q}$ and $\mathcal{P}_u \leftarrow \tilde{\mathcal{P}}$
        \State Add $u$ to $\operatorname{Ch}(v)$ and push $u$ to $\textsc{Queue}$
      \EndIf
    \EndFor
  \EndWhile
  \Statex \textit{Stage 2: bottom-up co-generation and revision}
  \State $(x, \mathcal{R}(T)) \leftarrow \mathcal{M}_{\mathrm{synth}}(T)$
  \State $(x, \mathcal{R}(T)) \leftarrow \mathcal{M}_{\mathrm{verify}}(x, T, \mathcal{R}(T))$
  \State $\mathcal{D} \leftarrow \mathcal{D} \cup \{(x, T, \mathcal{R}(T))\}$

\EndFor
\end{algorithmic}
\vspace{0.35em}
\hrule
\end{minipage}
\end{figure*}

\begin{enumerate}[nosep,leftmargin=*]
\item \textbf{Seed topic sampling.}  We sample diverse seed topics from both corpora.

\item \textbf{Recursive decomposition.}  Starting from a seed topic, we use DeepSeek-V3.2~\citep{deepseek_v3} to recursively decompose the topic into sub-queries.  At each internal node, the model generates 2--6 child sub-queries that collectively cover distinct parallel facets of the parent (e.g., mechanisms, stakeholders, risks, quantitative evidence, alternatives).  The branching factor is depth-aware: 6 children at the root, 4 at intermediate levels, and 3 at deeper levels, up to the configured maximum depth.  The decomposition prompt (Figure~\ref{fig:decompose_prompt}) explicitly requires coverage of comparison/alternatives, risks/failures, and quantitative evidence.

\item \textbf{Gating.} At each level, a separate LLM call determines whether further branching would add meaningful evidence.  Expansion continues only if the proposed children open substantively different facets.

\item \textbf{Evidence retrieval and extraction.}  At each node, we retrieve supporting passages from the source corpus using the node's sub-query.  An evidence extractor (Figure~\ref{fig:evidence_extract_prompt}) produces 1--3 concise factual statements per passage, grounding each node in specific retrieved content.

\item \textbf{Termination.}  Expansion terminates when a sub-query is judged atomic (i.e., further branching is unlikely to add novel, decision-relevant evidence) or when it reaches the configured maximum depth.  The resulting trees average 54.48 nodes and 38.66 leaves, providing a broad evidence set for downstream synthesis.

\item \textbf{Bottom-up merge and rubric generation.}  The complete tree (nodes + leaf evidence) is serialized into a compact JSON representation and fed to DeepSeek-V3.2 with the bottom-up merge prompt (Figure~\ref{fig:merge_prompt}).  The model (a)~selects a small set of leaf nodes covering distinct facets, (b)~iteratively merges them bottom-up through intermediate queries into a single final research question, and (c)~generates grounded rubric criteria, each with a type (factual/logical), description, importance weight in $[0, 1]$, source leaf IDs, and supporting evidence for factual rubrics.

\item \textbf{Verification and revision.}  A separate verification pass using GPT-5.1~\citep{gpt5} audits each sample on six dimensions: question--rubric alignment, deep-research suitability, rubric quality (atomic, non-redundant), evidence sufficiency, merge faithfulness, and scoring validity.  The auditor returns a KEEP/REVISE/DROP decision; REVISE samples receive repaired questions and rubrics.  This stage conservatively revises 91.50\% of generated tuples and drops 7.87\%.
\end{enumerate}

We retain 9{,}064 KEEP or REVISE examples and use all retained pairs for training.

\paragraph{Tree-depth upper bound.}
We set the maximum node depth to 3 to control construction cost and avoid excessive expansion.  The purpose of the tree is to provide enough complementary evidence facets for query--rubric co-generation, not to exhaustively enumerate every possible subtopic.  Under this upper bound, the generated trees already contain four evidence levels, 54.48 nodes, 38.66 leaves, and 2.54 evidence passages per leaf on average; 91.40\% of terminal leaves reach depth 3.  These statistics indicate that coverage is already high for our intended use: each tree provides dozens of candidate evidence leaves, from which the bottom-up merge selects the subset most relevant to a coherent research question and rubric.  Increasing the upper bound would also increase retrieval calls, serialized context length, duplicate or weakly related leaves, and topic drift, which can make the final question and rubrics harder to verify.  We therefore use depth 3 as a cost-controlled upper bound that provides sufficient multi-facet coverage while keeping synthesis, audit, and revision reliable.

\begin{center}
\begin{promptbox}[title=Recursive Decomposition Prompt]
\footnotesize\ttfamily
You are expanding a research search tree.\\
Parent query: ``\{parent\_query\}''\\
Current depth: \{depth\}, max depth: \{max\_depth\}.\\[4pt]
Current evidence (snippets + extracted statements):\\
\{snippets\} \\
Statements (sample, deduped top 8): \{statements\}\\[4pt]
\textbf{Goal}: maximize coverage AND depth potential. Propose up to \{limit\} sibling queries that:\\
\hangindent=1em -- Each covers a \textbf{different parallel facet} of the parent: stakeholders, timeline, mechanisms/causality, outcomes, risks/failures, alternatives, quantitative data, geography, policy, controversy.\\
\hangindent=1em -- Ensure at least one query targets \textbf{comparison/alternatives}, one targets \textbf{risks/failures}, and one targets \textbf{quantitative evidence}.\\
\hangindent=1em -- Keep at the same abstraction level as the parent.\\
\hangindent=1em -- Avoid overlap; prioritize facets not yet evidenced.\\
\hangindent=1em -- Keep queries concise (<12 tokens), factual, non-redundant.\\[4pt]
Return JSON: \texttt{\{"queries": ["q1", "q2", "q3"]\}}
\end{promptbox}
\captionof{figure}{\textbf{Recursive decomposition prompt.}  Branching factor is depth-aware: 6 children at the root, 4 at intermediate levels, 3 at deeper levels.}
\label{fig:decompose_prompt}
\end{center}

\begin{center}
\begin{promptbox}[title=Evidence Extraction Prompt]
\footnotesize\ttfamily
You are an evidence extractor. Given search queries, an action note (goal), and a webpage summary, return 1--3 concise factual statements for each query from the summary that are relevant to the queries and help achieve the action note.\\[4pt]
Queries: \texttt{\{queries\}}\\
Action note: \texttt{\{note\}}\\
Webpage summary: \texttt{\{summary\}}\\[4pt]
Return JSON: \texttt{\{"statements": ["...", "..."]\}}
\end{promptbox}
\captionof{figure}{\textbf{Evidence extraction prompt.}  Applied at every tree node after retrieval to extract atomic factual statements from passages.}
\label{fig:evidence_extract_prompt}
\end{center}

\begin{figure*}[t]
\centering
\begin{promptbox}[title=Bottom-Up Merge \& Rubric Generation Prompt]
\footnotesize\ttfamily
You are an autonomous agent for proposing \textbf{deep-research questions} from a search tree.\\[4pt]
Input is a compact JSON representation of a search tree with: \texttt{nodes} (id, parent, depth, query, children) and \texttt{leaves} (leaf id, path root$\to$leaf, evidence statements).\\[4pt]
Your task is to define a \textbf{complete, well-formed research problem} that ultimately requires a long-form, structured answer. You are expected to make the question self-contained and answerable in scope (but you are NOT expected to solve it).\\[4pt]
Important: The tree may contain \textbf{noise} or even harmful/irrelevant information not related to the root topic. You must identify and avoid such nodes/claims when forming the question and rubrics. If a leaf is misleading but must be addressed, reflect this with \textbf{negative-weight rubrics} that require debunking or exclusion.\\[4pt]
\textbf{You must:}\\
\hangindent=1em 1) Select a \textbf{small set of leaf nodes} that best cover distinct facets of the root topic.\\
\hangindent=1em 2) Perform a \textbf{bottom-up merge}: combine selected leaf queries into intermediate queries, then merge those into the final research question.\\
\hangindent=1em 3) Create rubrics \textbf{grounded only in statements from the selected leaves}.\\
\hangindent=1em 4) Provide a \textbf{visualizable reasoning trace}: which leaves were selected, why, and how they were merged step-by-step.\\[6pt]
\textbf{Bottom-up merge requirements:}\\
\hangindent=1em -- Start from selected leaves (last-layer nodes). Group them into \textbf{multiple} intermediate queries.\\
\hangindent=1em -- Merge intermediate queries \textbf{iteratively} into higher-level queries until one final question.\\
\hangindent=1em -- The merge can take many steps; include it in \texttt{merge\_trace} with explicit inputs, outputs, and short rationale. Each merge step should explain why those inputs were grouped and what new query they form.\\[6pt]
\textbf{Rubric Types} (each rubric MUST belong to exactly one):\\
\hangindent=1em -- \textbf{Logical}: structure/organization/reasoning checks for a long-form answer.\\
\hangindent=1em -- \textbf{Factual}: specific facts/claims that must be addressed, each supported by evidence from selected leaves only.\\[6pt]
\textbf{Weighting:}\\
\hangindent=1em -- Importance score between 0 and 1; higher = more critical.\\
\hangindent=1em -- Weights do not need to sum to 1.\\[6pt]
\textbf{Rubric quality (applies to every rubric):}\\
\hangindent=1em -- Make each rubric precise, atomic, and binary-checkable; avoid vague verbs.\\
\hangindent=1em -- Anchor factual rubrics to concrete supporting statements from selected leaf evidence; do not invent new facts.\\
\hangindent=1em -- Cover evidence breadth \& attribution, reasoning soundness, completeness, organization/clarity, and handling of ambiguity or counterpoints.\\
\hangindent=1em -- Keep rubrics concise and avoid duplicate checks.\\
\hangindent=1em -- Each rubric must be evaluated independently (no dependencies between rubrics).\\
\hangindent=1em -- Every rubric MUST include \texttt{source\_leaf\_ids} $\subseteq$ \texttt{selected\_leaf\_ids}.\\[6pt]
\textbf{Question Construction:}\\
\hangindent=1em -- Motivated by selected rubrics and evidence; expose meaningful gaps or unresolved relationships.\\
\hangindent=1em -- Must NOT be fully answerable from provided material alone.\\
\hangindent=1em -- Do NOT list rubrics or solution steps in the question.\\
\hangindent=1em -- Brief, precise, research-heavy. Do NOT invent facts beyond the material.\\
\hangindent=1em -- Minimum 5 rubrics (numbered R1, R2, R3, ...).\\[6pt]
Return JSON only: \texttt{\{question, selected\_leaf\_ids, selection\_trace, merge\_trace, rubrics\}}.
\end{promptbox}
\captionof{figure}{\textbf{Bottom-up merge and rubric generation prompt.} This prompt is the core of the evidence-tree pipeline, converting raw tree evidence into a research question paired with grounded rubric criteria. The \texttt{source\_leaf\_ids} $\subseteq$ \texttt{selected\_leaf\_ids} constraint ensures all rubrics remain grounded in retrieved evidence.}
\label{fig:merge_prompt}
\end{figure*}

\subsection{Data Statistics}
\label{app:data_stats}

Table~\ref{tab:data_stats} summarizes the statistics of the generated training data.

\begin{table}[t]
\small
\centering
\begin{tabular}{lc}
\toprule
\textbf{Statistic} & \textbf{Value} \\
\midrule
Generated examples & 9{,}838 \\
Final retained (KEEP + REVISE) & 9{,}064 (92.1\%) \\
Final training pairs & 9{,}064 \\
\midrule
Tree depth (fixed) & 3 \\
Nodes per tree & 54.5 \\
Leaves per tree & 38.7 \\
Selected leaves per example & 5.0 \\
Evidence passages per selected leaf & 2.9 \\
\midrule
Rubric criteria per example & 7.0 \\
\quad Factual / Logical & 58.4\% / 41.6\% \\
Revision rate & 91.5\% \\
Drop rate & 7.9\% \\
\midrule
Tree generation model & DeepSeek-V3.2 \\
Quality audit model & GPT-5.1 \\
SFT subset size & 200 \\
SFT annotation model & GPT-5.1 \\
\bottomrule
\end{tabular}
\caption{\textbf{Training data statistics.} Revision indicates auditor-guided repair rather than rejection; retained examples include both KEEP and REVISE cases.}
\label{tab:data_stats}
\end{table}

\subsection{Supervision Pool Scaling}
\label{app:data_scaling}

\begin{table}[t]
\centering
\small
\resizebox{\columnwidth}{!}{
\renewcommand{\arraystretch}{1.2}
\setlength{\tabcolsep}{4pt}
\begin{tabular}{l l c c c | c c}
\toprule
\textbf{Pool} & \textbf{Budget} & SQAv2 & ResearchQA & DRB & Avg. & $\Delta$ \\
\midrule
\rowcolor{oursRow}
9K & 140 steps & 86.0 & 75.2 & 43.6 & 68.3 & \\
9K & 285 steps & 84.5 & 75.9 & 43.8 & 68.1 & $-$0.2 \\
\midrule
18K & 140 steps & 85.4 & 75.8 & 43.1 & 68.1 & \\
\rowcolor{oursRow}
18K & 280 steps & 86.3 & 78.0 & 43.9 & \textbf{69.4} & \textbf{+1.3} \\
\bottomrule
\end{tabular}
}
\caption{\textbf{Scaling the supervision pool.} $\Delta$ is the change from the 140-step to the ${\sim}$280-step checkpoint within each pool. The two pools are comparable after one pass over 9K, but extending training helps only with 18K: the plateau is a property of the smaller pool, not of the method or the bounded corpora.}
\label{tab:data_scaling}
\end{table}

The plateau in \S\ref{sec:training_efficiency} raises a question the 9K setting alone cannot answer: does it reflect a limit of the method and its bounded environment, or the finite diversity of one particular pool?
To separate these, we ran the pipeline over additional seed topics from the same corpora to obtain ${\sim}$18K independently synthesized pairs, and trained to 280 steps under an otherwise identical recipe (Table~\ref{tab:data_scaling}).

The comparison is an interaction rather than a level effect: at 140 steps---roughly one pass over the 9K pool---the two pools are comparable (68.3 vs.\ 68.1), so doubling the data buys nothing on its own; the two diverge only once training continues past one pass.
Extending to ${\sim}$280 steps costs 0.2 points with 9K but gains 1.3 with 18K, and the flat segment of the \textsc{DeepRubric} curve in Figure~\ref{fig:step_efficiency} after 140 steps is what the larger pool removes: SQAv2 falls from 86.0 to 84.5 with 9K but recovers to 86.3 with 18K, and the best 18K checkpoint is 1.1 above the best 9K one.
The largest gain is on ResearchQA (78.0 vs.\ 75.9), consistent with what a larger pool mainly adds---query diversity across research fields.
The earlier plateau is therefore a property of the finite 9K pool rather than of evidence-tree supervision or the bounded corpora.

\paragraph{Construction of the expanded pool.}
We ran the same pipeline over an additional, disjoint set of seed topics sampled from the same two corpora, producing approximately 9K further query--rubric pairs for a combined pool of ${\sim}$18K.
All construction settings are unchanged: identical tree depth and branching bounds, the same DeepSeek-V3.2 expansion and synthesis stages, the same GPT-5.1 verification pass, and comparable retention and per-pair statistics.
The expansion therefore adds new supervision rather than re-sampling or augmenting the original 9K pairs, which is what makes the comparison in Table~\ref{tab:data_scaling} a test of pool size rather than of data reuse.
Training on the 18K pool uses the same GRPO recipe as the main run (Table~\ref{tab:grpo_hypers}); at 512 rollouts per step, 280 steps again correspond to roughly one pass over the pool.

\subsection{Verification Statistics and Strong-Model Influence}
\label{app:verify_stats}

A natural concern is that the 91.5\% revision rate might indicate that the final supervision is generated primarily by the strong verifier rather than by the evidence tree. We clarify that the verifier is not used as an open-ended content generator. As shown in Table~\ref{tab:revision_reasons}, the dominant revision categories are structural alignment errors---question--rubric mismatch, evidence sufficiency, scope or merge overreach, invalid weights, and rubric granularity---rather than the addition of new factual targets.

The verification prompt (Figure~\ref{fig:verify_prompt}) further enforces tree-grounded revision: revised factual rubrics must cite \texttt{source\_leaf\_ids} that are a subset of the already selected tree leaves, and their supporting evidence must be verbatim or tightly faithful paraphrases of selected-leaf statements. Thus, the verifier cannot introduce external evidence or new factual support; it can only repair how tree-derived information needs are expressed as a query and evaluation criteria.

We therefore view the verifier as improving the \emph{quality and alignment} of the reward specification, rather than changing the factual \emph{source} of the supervision. This distinction is also supported by the no-revision ablation: removing the verifier lowers performance, but the unrevised tree-based rubrics still outperform query-first rubric baselines, indicating that the evidence tree itself provides substantial supervision signal.

\begin{table}[t]
\small
\centering
\begin{tabular}{lr}
\toprule
\textbf{Primary Revision Reason} & \textbf{Rate} \\
\midrule
Question--rubric mismatch & 56.4\% \\
Evidence sufficiency / grounding & 18.2\% \\
Question scope / merge overreach & 8.3\% \\
Negative-weight rubric removal & 8.2\% \\
Rubric design / granularity & 5.3\% \\
Entity / domain / time mismatch & 2.5\% \\
Other & 1.0\% \\
\bottomrule
\end{tabular}
\caption{\textbf{Primary revision reasons.} All categories are structural alignment corrections; none involves generating new factual content.}
\label{tab:revision_reasons}
\end{table}

\begin{figure*}[t]
\centering
\begin{promptbox}[title=Quality Verification System Prompt]
\footnotesize\ttfamily
You are an expert auditor for tree-rubric data used in long-form answer evaluation and deep research QA. You must audit one JSON sample.\\[4pt]
\textbf{Goal}: Decide if the sample should be KEEP / REVISE / DROP. If REVISE, provide a repaired question and/or repaired rubrics. Keep the original topic unless structurally unsalvageable.\\[4pt]
\textbf{Deep research standards}: Requires synthesis across multiple evidence sources; multi-step reasoning, not shallow fact lookup; connects mechanisms, evidence, limitations, uncertainty, and conclusions; rewards structured, evidence-grounded comparative reasoning.\\[4pt]
\textbf{Audit dimensions} (apply all, do not skip):\\
\hangindent=1em 1) \textbf{Question--rubric alignment}: Rubrics must stay centered on the main question. Granularity should match. Do not elevate a narrow technical detail into a mandatory rubric unless truly central.\\
\hangindent=1em 2) \textbf{Deep research suitability}: Rubrics should evaluate deep reasoning---comparison, synthesis, causal/mechanistic explanation, evidence qualification, structured conclusions---not reduce to a checklist of isolated facts.\\
\hangindent=1em 3) \textbf{Rubric quality}: Each rubric atomic, specific, independently judgeable, non-redundant. Avoid vague or non-judgeable wording.\\
\hangindent=1em 4) \textbf{Evidence sufficiency}: \texttt{source\_leaf\_ids} must sufficiently support each rubric. A detail in one leaf does not automatically justify a required rubric. Flag claims stronger than cited support. Note: leaf nodes may be truncated; judge based on visible information.\\
\hangindent=1em 5) \textbf{Merge faithfulness}: Final question should reflect semantic union of selected leaves and intermediate merges. Flag unjustified broadening, strengthening, or narrowing.\\
\hangindent=1em 6) \textbf{Scoring validity}: Weights reasonable; core factual rubrics not dominated by vague logical ones; negative-weight rubrics treated as problematic unless explicitly justified.\\[4pt]
\textbf{Decision policy}: KEEP (usable as-is); REVISE (valuable but needs targeted repair); DROP (severe structural or semantic failure). Prefer REVISE over DROP when topic remains valuable. Do not invent unsupported claims.\\[4pt]
\textbf{If REVISE, follow hard rewrite template}:\\
\hangindent=1em -- \textit{revised\_question}: Keep topic, fix overreach/under-specification. Must stay faithful to selected leaves + merge trace. Must remain a deep research question.\\
\hangindent=1em -- \textit{revised\_rubrics}: Output 5--8 rubric objects. Every rubric MUST contain: id, type (logical/factual), description, weight, \texttt{source\_leaf\_ids}. \texttt{source\_leaf\_ids} must be $\subseteq$ \texttt{selected\_leaf\_ids}. Logical rubrics do NOT need evidence. Factual rubrics MUST include non-empty evidence grounded in selected-leaf statements (verbatim or tightly faithful short paraphrase). Avoid semantic duplicates. Prefer rubrics evaluating synthesis, evidence-based comparison, causal reasoning, limitations, and conclusion quality.\\[4pt]
Return valid JSON only: \texttt{\{decision, error\_tags, major\_issues, justification, fix\_plan, revised\_question, revised\_rubrics, implementation\_guidance\}}.
\end{promptbox}
\captionof{figure}{\textbf{Quality verification system prompt.} The key constraint is that all revised rubrics must ground \texttt{source\_leaf\_ids} within the tree's \texttt{selected\_leaf\_ids} and provide evidence from leaf statements only, preventing the verifier from injecting parametric knowledge.}
\label{fig:verify_prompt}
\end{figure*}

\section{Tool and Inference Details}
\label{app:inference_details}

During both training and evaluation, the agent uses the same three retrieval-tool interfaces, but the backends differ between local RL rollouts and live benchmark inference:

\begin{itemize}[nosep,leftmargin=*]
\item \textbf{\texttt{search}}: Issues a keyword query and returns ranked snippets (dense retrieval over the local Wikipedia/OpenScholar corpora during training; Google Search via Serper API during evaluation).

\item \textbf{\texttt{browse}}: Given a URL or document identifier, retrieves and returns the full text of the page (direct corpus lookup during training; Jina Reader during evaluation).

\item \textbf{\texttt{scholar}}: Issues a scientific-literature query and returns paper snippets (dense retrieval over OpenScholar during training; Google Scholar via Serper API during evaluation).
\end{itemize}

Tool calls are issued via structured \texttt{<tool\_call>} JSON blocks within the generation.
Each tool call includes a \texttt{name} field and an \texttt{arguments} field (containing the query or URL).
Observations are returned in \texttt{<tool\_response>} blocks and masked during gradient computation to avoid training on retrieval artifacts.
We cap the number of tool call turns at 20 during training and 10 during evaluation to balance efficiency and performance.

\subsection{System Prompt}
\label{app:system_prompt}

We use a unified system prompt for both training and evaluation.
The prompt instructs the model to act as a deep research assistant that performs iterative, multi-source investigations.
Key elements include: (1)~mandatory multi-round search (at least 2--4 rounds of tool calls), (2)~structured output with \texttt{<think>}, \texttt{<tool\_call>}, and \texttt{<answer>} tags, (3)~inline citation requirements using \texttt{<cite id="...">...</cite>} grounded only in returned snippets, and (4)~a strict workflow alternating reasoning $\to$ tool call / answer.
The full system prompt is shown in Figure~\ref{fig:system_prompt}.

\begin{figure*}[t]
\centering
\begin{systempromptbox}[title=Deep Research Agent System Prompt (Part 1: Task \& Answer Requirements)]
\footnotesize\ttfamily
You are a deep research assistant. Your core function is to conduct thorough, multi-source investigations into any topic. You must handle both broad, open-domain inquiries and queries within specialized academic fields.\\[4pt]
For every request, you MUST perform iterative, multi-step information gathering. You should search as broadly and deeply as possible, gather as much relevant information as is reasonably available, and synthesize evidence from credible, diverse sources to deliver a comprehensive, accurate, and objective response.\\[4pt]
You MUST NOT rely on a single search. Instead:\\
\hangindent=1em -- Always perform multiple rounds of search using different queries, perspectives, or keyword variations.\\
\hangindent=1em -- After each search, analyze gaps, uncertainties, or missing aspects, and issue follow-up searches.\\
\hangindent=1em -- Continue searching until the key aspects of the question are sufficiently covered.\\
\hangindent=1em -- You should typically perform at least 2--4 rounds of tool calls before producing the final answer, unless the question is extremely simple.\\[4pt]
A response is NOT sufficient if: only one source or one perspective is used; key concepts in the question are not individually investigated; there is no cross-source verification.\\[4pt]
The final answer MUST be comprehensive, detailed, and in-depth. It should fully address all aspects of the question, explain underlying mechanisms, compare different perspectives, and provide clear reasoning supported by evidence.\\[4pt]
When you have gathered sufficient information and are ready to provide the definitive response, you must enclose the entire final answer within \texttt{<answer></answer>} tags.\\[4pt]
You should ground every nontrivial claim in retrieved snippets. Cite using \texttt{<cite id="...">...</cite>} drawn only from returned snippets. Prefer authoritative sources (peer-reviewed papers, reputable benchmarks/docs) and prioritize recent work for fast-moving areas. Acknowledge uncertainty and conflicts; if evidence is thin or sources disagree, state it and explain what additional evidence would resolve it.\\[4pt]
Structure the answer with clear markdown headers and a coherent flow. In each section, write 5--8 sentence paragraphs with clear topic sentences and transitions.\\[4pt]
The answer MUST: be comprehensive and cover all key aspects; provide detailed explanations rather than brief summaries; compare different models, assumptions, or perspectives when relevant; explain causal mechanisms and not just describe phenomena; synthesize information across sources into a coherent narrative.\\[4pt]
Use lists sparingly only when they improve clarity. Synthesize rather than enumerate content: group findings across papers, explain relationships, and build a coherent narrative supported by citations.\\[4pt]
DO NOT invent snippets or citations and never fabricate content.\\[4pt]
You can reason and use tools iteratively, but all reasoning text must appear only inside a standalone \texttt{<think>...</think>} block, and all tool calls must appear only inside a standalone \texttt{<tool\_call>...</tool\_call>} block. Tool calls must never appear inside a \texttt{<think>} block.
\end{systempromptbox}
\captionof{figure}{\textbf{System prompt for the deep research agent (Part 1).} Task definition, search requirements, citation policy, and answer formatting rules.}
\label{fig:system_prompt}
\end{figure*}

\begin{figure*}[t]
\centering
\begin{systempromptbox}[title=Deep Research Agent System Prompt (Part 2: Tools \& Workflow)]
\footnotesize\ttfamily
\textbf{Tool Usage Constraints}\\[2pt]
\hangindent=1em -- The \texttt{browse} tool can ONLY be used on URLs returned by a previous \texttt{search} call.\\
\hangindent=1em -- You MUST NOT fabricate, guess, or manually construct URLs for browsing.\\
\hangindent=1em -- If additional webpages are needed, first use \texttt{search} to retrieve them, then select from returned results.\\
\hangindent=1em -- Prioritize browsing multiple distinct URLs from different sources to ensure diversity of evidence.\\[6pt]
\textbf{Available Tools}\\[2pt]
\texttt{\{"name":"search","description":"Perform web searches and return top results.","parameters":\{"query":[list of strings]\}\}}\\[2pt]
\texttt{\{"name":"browse","description":"Open a specific URL and return readable content.","parameters":\{"url":"string"\}\}}\\[2pt]
\texttt{\{"name":"scholar","description":"Retrieve information from scientific papers.","parameters":\{"query":[list of strings]\}\}}\\[6pt]
\textbf{Tool Call Format}\\[2pt]
\texttt{<tool\_call>\{"name":"tool\_name","arguments":\{...\}\}</tool\_call>}\\[6pt]
\textbf{Tool Output Format}\\[2pt]
For search/scholar: \texttt{<tool\_response><snippet id=ID>content</snippet>...</tool\_response>}\\
For browse: \texttt{<tool\_response><webpage id=ID>content</webpage></tool\_response>}\\[4pt]
Support every non-trivial claim with retrieved evidence. Wrap the exact claim span in \texttt{<cite id="ID1,ID2">...</cite>}, where IDs are snippet IDs from returned results (comma-separated if multiple). Use only returned snippets; never invent IDs.\\[6pt]
\textbf{Workflow Rules}\\[2pt]
After every \texttt{</tool\_response>}, the assistant must first output a standalone \texttt{<think>...</think>} block. Then it must output either: (1) exactly one standalone \texttt{<tool\_call>} block if more evidence is needed, or (2) exactly one standalone \texttt{<answer></answer>} block if evidence is sufficient.\\[6pt]
\textbf{Final Answer Rules}\\[2pt]
\hangindent=1em -- Generate a comprehensive, detailed, and in-depth final answer marked with \texttt{<answer></answer>}.\\
\hangindent=1em -- Avoid short or shallow responses; prioritize depth, clarity, and completeness.\\
\hangindent=1em -- Wrap supported text in \texttt{<cite id="SNIPPET\_ID">...</cite>} using exact IDs from returned results.\\
\hangindent=1em -- If multiple sources support a passage, use multiple \texttt{<cite>} tags around relevant clauses.
\end{systempromptbox}
\captionof{figure}{\textbf{System prompt for the deep research agent (Part 2).} Tool definitions, call/output format, workflow rules, and final answer requirements. The same prompt is used for both training and evaluation; during training the tools connect to local corpora, during evaluation they connect to live APIs.}
\label{fig:system_prompt_2}
\end{figure*}

\section{Evaluation Details}
\label{app:eval_details}

\subsection{Benchmark Descriptions}
\label{app:benchmarks}

We provide additional details on the three evaluation benchmarks:

\begin{itemize}[nosep,leftmargin=*]
\item \textbf{AstaBench-ScholarQA-CS2 (SQAv2)}~\citep{sqav2}: We use the official evaluation code to evaluate all methods on the 100 test-set questions. We compute rubric score, answer precision, citation precision, and citation recall using Gemini-2.5-Flash as the judge, following~\citet{sqav2}.

\item \textbf{ResearchQA}~\citep{researchqa}: We evaluate using the official ResearchQA evaluation suite with GPT-4.1-mini as the judge. We compute averaged rubric scores on a 200-question subset.

\item \textbf{DeepResearch Bench (DRB)}~\citep{drb}: We use the official evaluation code on 50 English open-ended deep research questions. We compute Comprehensiveness, Insight/Depth, Instruction-Following, and Readability, and report Overall as the macro average of component metrics. We use Gemini-2.5-Pro as the judge following~\citet{drb}.
\end{itemize}

\subsection{Decomposing the Efficiency Gain}
\label{app:efficiency_analysis}

The ${\sim}$13$\times$ RL GPU-hour gap between \textsc{DeepRubric} and DR~Tulu (750 vs.\ 9{,}700, Table~\ref{tab:training_cost}) combines two distinct factors: how much GPU time each rollout costs, and how many rollouts are needed to reach a given score.
We separate them here because only the second is a property of the supervision.

The first factor is environmental.
Our RL rollouts retrieve from local corpora, which removes live-API latency from the critical path, and the two runs use different accelerators (8$\times$A100 for ours, 8--16$\times$H100 for DR~Tulu).
Both effects reduce wall-clock and GPU time per rollout independently of what the rubrics contain, so we do not attribute this component to evidence-tree supervision.

The second factor is independent of both latency and hardware.
\textsc{DeepRubric} reaches 68.3 average in 140 GRPO steps, while DR~Tulu reaches 68.2 in 1{,}900 steps.
At their respective per-step batches (512 trajectories for ours, 256 for DR~Tulu), this is 13.6$\times$ fewer policy updates and 6.8$\times$ fewer rollouts in total (71{,}680 vs.\ 486{,}400).
The gap appears early rather than only at convergence: our 75-step checkpoint reaches 67.1 after 38{,}400 rollouts, already exceeding DR~Tulu's 66.5 at 1{,}000 steps (256{,}000 rollouts).
Rollout counts are unaffected by retrieval latency or GPU model, so cheaper rollouts cannot explain why fewer of them suffice.

Two further observations support reading this second component as a property of the construction.
The ablations in \S\ref{sec:ablation} hold the retrieval environment, training queries, and step budget fixed and vary only how rubrics are built, and the gap persists there.
\textsc{DeepRubric} and DR~Tulu do differ in more than rubric construction, but the diverse query distribution (\S\ref{sec:generated_data_analysis}) and the bounded local corpus are themselves products of the proposed pipeline rather than external confounds, so we read the rollout-efficiency component as attributable to the construction as a whole rather than to any single isolated factor.

\subsection{DRB Results by Task Category}
\label{app:drb_category}

DRB groups its questions into expert research categories, which allows a finer-grained check of whether gains are broad or concentrated in a few topics.
Table~\ref{tab:drb_category} reports the categories containing at least four questions; categories with fewer questions are omitted because per-category averages become unstable.
\textsc{DeepRubric}-8B improves over DR~Tulu-8B on every category shown, including domains far from the scholarly and encyclopedic content of our training corpora, such as finance and business.
Per-category sample sizes are small, so individual margins should not be over-interpreted; the relevant observation is the consistency of the direction.

\begin{table}[t]
\small
\centering
\begin{tabular}{lccc}
\toprule
\textbf{DRB category} & \textbf{Ours} & \textbf{DR Tulu} & $\Delta$ \\
\midrule
Finance \& Business & \textbf{45.14} & 41.93 & +3.21 \\
Software Development & \textbf{43.40} & 41.32 & +2.08 \\
Health & \textbf{43.11} & 42.10 & +1.01 \\
Education \& Jobs & \textbf{44.86} & 44.35 & +0.51 \\
\bottomrule
\end{tabular}
\caption{\textbf{DRB overall scores by task category}, for categories with at least four questions. Both models are the 8B checkpoints from Table~\ref{tab:main_results}.}
\label{tab:drb_category}
\end{table}

\subsection{Human Evaluation Protocol}
\label{app:human_eval}

Because both our reward and our automatic evaluation involve LLM judges, one may ask whether the reported gains reflect adaptation to a particular judge rather than better reports.
Evaluation is separated from training by construction: the three benchmarks are held out from the entire pipeline---their queries, rubrics, and reference materials are never used to synthesize data, compute rewards, or update the model---their criteria are independently human-written, expert-crafted, or human-validated, and their official judges (Gemini-2.5-Flash, GPT-4.1-mini, Gemini-2.5-Pro) exclude our training reward judge (Qwen3.5-35B-A3B).
Gains therefore have to transfer across three independent protocols and a different judge family from the one optimized against.
As a further check that does not rely on any LLM judge, we conducted a blinded pairwise human comparison between \textsc{DeepRubric}-8B and DR~Tulu-8B, the closest matched baseline.

\paragraph{Setup.}
We sampled 30 evaluation questions and collected one report from each system per question, giving 30 pairs.
Each pair was rated independently by three evaluators with research experience.
Reports were anonymized, and the A/B presentation order was counterbalanced across pairs so that neither system consistently appeared first.
Evaluators chose the better report or recorded a tie when the two were of comparable quality, along three dimensions: \emph{factual and evidential support} (are claims accurate and backed by the cited sources?), \emph{citation quality} (are citations present, relevant, and correctly attributed?), and \emph{overall preference}.

\paragraph{Results.}
Table~\ref{tab:human_eval} reports win/tie/loss counts by majority vote across the three evaluators.
\textsc{DeepRubric} is preferred on all three dimensions, with the clearest margin on factual and evidential support---the dimension most directly targeted by evidence-grounded rubric supervision.
The sample is small and the margins are correspondingly modest, so we read this as a directional consistency check on the automatic evaluation rather than as an independent effect-size estimate.

\begin{table}[t]
\small
\centering
\begin{tabular}{lccc}
\toprule
\textbf{Evaluation dimension} & \textbf{Win} & \textbf{Tie} & \textbf{Loss} \\
\midrule
Factual / evidential support & 15 & 8 & 7 \\
Citation quality & 11 & 12 & 7 \\
Overall preference & 12 & 10 & 8 \\
\bottomrule
\end{tabular}
\caption{\textbf{Blinded human comparison} of \textsc{DeepRubric}-8B against DR~Tulu-8B over 30 report pairs, three evaluators per pair. Win/loss are from the perspective of \textsc{DeepRubric}.}
\label{tab:human_eval}
\end{table}

\subsection{Pairwise Factual-Discrimination Evaluation}
\label{app:pairacc}

This appendix details the PairAcc evaluation reported in \S\ref{sec:pairacc}, which measures rubric quality on the synthesized data directly rather than through a training run.

\paragraph{Constructing response pairs.}
We sampled 750 queries from the synthesized supervision pool and generated one report per query with \textsc{DeepRubric}-8B.
For each report we produced a corrupted counterpart by replacing 3--5 factual claims with plausible but incorrect alternatives, targeting four claim types that are locally checkable and unambiguous to corrupt: named entities, dates, numeric years, and negations (for example, flipping ``was not replicated'' to ``was replicated'').
Corruptions preserve surface fluency, sentence structure, and overall topic, so the original and corrupted reports differ only in factual correctness and are otherwise comparable in length and organization.
No corruption targets the rubric text itself.

\paragraph{Scoring.}
Each original--corrupted pair is scored twice under an identical judging protocol, prompt template, and judge model, differing only in which rubric set is supplied: (i) the evidence-tree rubrics synthesized for that query, or (ii) search-based (query-first) rubrics generated for the same query by retrieving passages with the query and then producing criteria from the retrieved context---the same construction used for the \emph{Search-based rubrics} ablation in \S\ref{sec:ablation}.
Both rubric sets therefore evaluate the same 750 pairs under the same conditions.

\paragraph{Metric.}
A pair counts as correct when the rubric assigns a strictly higher factual score to the uncorrupted original; ties count as 0.5.
PairAcc is the mean over the 750 pairs.
Confidence intervals are computed by paired bootstrap over pairs with 10{,}000 resamples, which respects the fact that both rubric sets are evaluated on identical items.
Evidence-tree rubrics reach 93.87 and search-based rubrics 86.80, a difference of 7.07 points with a 95\% CI of [5.20, 8.93].
Equivalently, the fraction of corruptions that the rubric fails to detect falls from 13.20\% to 6.13\%.

\subsection{Details of Baselines}
\label{app:baselines}

For all open deep research baselines, we run their officially released code with default or recommended configurations:
\begin{itemize}[nosep,leftmargin=*]
\item \textbf{Search-R1-7B}~\citep{searchr1}: We use the official checkpoint and inference code.
\item \textbf{ASearcher-Web-7B}~\citep{asearcher}: We use the official checkpoint and inference code.
\item \textbf{WebExplorer-8B}~\citep{webexplorer}: We use the official checkpoint.  Note that WebExplorer does not release training code or data.
\item \textbf{WebThinker-32B-DPO}~\citep{webthinker}: We use the official DPO checkpoint.  For the fixed-pipeline variant (report mode), we additionally enable the iterative report generation module.
\item \textbf{Tongyi-DR-30B-A3B}~\citep{tongyi_dr}: We use the official checkpoint.
\item \textbf{DR~Tulu-8B}~\citep{drtulu_rler}: We use both the SFT and RL checkpoints with the official \texttt{dr-agent-lib} inference pipeline.
\item \textbf{Ai2 ScholarQA}~\citep{singh-etal-2025-ai2}: We use the official code with Claude Sonnet as the backbone LM.
\end{itemize}

For closed deep research systems (Perplexity Deep Research, Gemini~3~Pro~+~Search, GPT-5~+~Search, and OpenAI Deep Research), we use their public APIs with default settings.
Results marked with $^\dagger$ in Table~\ref{tab:main_results} are evaluated on a 100-sample subset due to cost constraints.
Results marked with $^*$ are reported by the original benchmark authors.

\section{Case Study Response Comparison}
\label{app:case_study}

Table~\ref{tab:case_study_final_response} shows the prompt, abridged final answers, and DRB scores for the case study discussed in \S\ref{sec:case_study}.
Ellipses omit intermediate details and citation markup.

\begin{table*}[t]
\centering
\footnotesize
\renewcommand{\arraystretch}{1.18}
\setlength{\tabcolsep}{5pt}
\begin{tabular}{p{0.13\textwidth}p{0.80\textwidth}}
\toprule
\textbf{Item} & \textbf{Content} \\
\midrule
\textbf{Question} &
Summarize global AI investments, key initiatives, and outputs by major international consulting firms, including the Big Four, Accenture, MBB, IBM, and Capgemini. The report should cover products/services, client cases, strategy, talent programs, and future trends. \\
\midrule
\textbf{DR~Tulu answer} &
\emph{``For this summary, `major international consulting firms' includes the Big Four \ldots MBB firms \ldots IBM \ldots Capgemini \ldots Accenture \ldots The focus is on their global investments, AI-driven products and services, client-facing initiatives \ldots strategic thrusts \ldots and talent development programs.''}

\emph{``Several high-impact items \ldots were not retrievable in this session and therefore are not asserted beyond cautious, sourced generalizations. A next-step due-diligence plan would capture \ldots annual reports \ldots firm newsrooms \ldots learning and development pages \ldots venture and research lab updates.''} \\
\midrule
\textbf{\textsc{DeepRubric} answer} &
\emph{``The available evidence points to a clear pattern: major international consulting firms are investing heavily in AI, but their approaches diverge by firm type and by stage of maturity.''}

\emph{``In conclusion, the major international consulting firms are pursuing AI through a mix of platform development, client-facing solutions, strategic advisory, and talent upskilling. Accenture and IBM are particularly prominent as platform builders \ldots Big Four are scaling AI at massive scale \ldots MBB firms are emphasizing strategy, governance, and human-capital transformation \ldots For all firms, the central challenge is translating AI ambition into sustained business value without losing sight of the human and organizational dimensions of transformation.''} \\
\midrule
\textbf{DRB score} &
\begin{tabular}{@{}lccccc@{}}
& Overall & Comp. & Depth/Insight & Instr. Follow. & Readability \\
DR~Tulu-8B & 32.6 & 24.7 & 33.8 & 37.5 & 37.2 \\
\textsc{DeepRubric}-8B & 47.6 & 47.1 & 47.6 & 54.2 & 38.5 \\
\end{tabular} \\
\bottomrule
\end{tabular}
\caption{\textbf{Abridged final-answer comparison for one DRB case study.} Ellipses mark omitted details.}
\label{tab:case_study_final_response}
\end{table*}

\end{document}